\documentclass[11pt,a4paper]{article}
\usepackage{linguex}
\usepackage[hyperref]{acl2020}
\usepackage{pslatex}
\usepackage{latexsym}
\usepackage{amsmath}
\usepackage{amssymb}
\usepackage{graphicx}
\graphicspath{ {./images/} }
\usepackage{svg}
\usepackage{booktabs}
\usepackage{tabularx}
\usepackage{xcolor}
\usepackage{dirtytalk}
\usepackage{fdsymbol}
\usepackage{hyperref}

\usepackage{amssymb,amsmath, graphicx}
\usepackage{tikz}

\definecolor{annotatedcolor}{RGB}{127, 127, 127}
\definecolor{surprisecolor}{RGB}{31, 119, 180}
\definecolor{suspensecolor}{RGB}{255, 127, 14}
\definecolor{surpriseentropycolor}{RGB}{44, 160, 44}
\definecolor{suspensestatecolor}{RGB}{214, 39, 40}

\usepackage{microtype}

\aclfinalcopy % Uncomment this line for the final submission
\title{Modelling Suspense in Short Stories as Uncertainty Reduction
  over Neural Representation}

\author{David Wilmot \and Frank Keller\\
  Institute for Language, Cognition and Computation \\
  School of Informatics, University of Edinburgh \\
  10 Crichton Street, Edinburgh EH8 9AB, UK \\
  \url{david.wilmot@ed.ac.uk},~~\url{keller@inf.ed.ac.uk}}

\date{}

\begin{document}

%% indentation for example sentences
\setlength{\Exlabelsep}{1ex}

\maketitle

\renewcommand{\UrlFont}{\ttfamily\small}

\begin{abstract}
  Suspense is a crucial ingredient of narrative fiction, engaging
readers and making
stories compelling. While there is a vast
  theoretical literature on suspense, it is computationally not well
  understood. We compare two ways for modelling suspense: surprise, a
  backward-looking measure of how unexpected the current state is
  given the story so far; and uncertainty reduction, a forward-looking
  measure of how unexpected the continuation of the story is.  Both
  can be computed either directly over story representations or over
  their probability distributions. We propose a hierarchical language
  model that encodes stories and computes surprise and uncertainty
  reduction. Evaluating against short stories annotated with human
  suspense judgements, we find that uncertainty reduction over
  representations is the best predictor, resulting in near human
  accuracy. We also show that uncertainty reduction can be used to
  predict suspenseful events in movie synopses.
\end{abstract}

%%%%%%%%%%%%%%%%%%%%%%%%%%%%%%%%%%%%%%%%%%%%%%%%%%%%%%%%%%%%%%%%%%%%%%%%%%%%%%%%
\section{Introduction}

As current NLP research expands to include longer, fictional texts, it
becomes increasingly important to understand narrative
structure. Previous work has analyzed narratives at the level of
characters and plot events \citep[e.g.,][]{gorinski-lapata-2018-whats,
  martin2018event}.  However, systems that process or generate
narrative texts also have to take into account what makes stories
compelling and enjoyable. We follow a literary tradition that makes
\textit{And then?} \citep{forster1985aspects,rabkin1973narrative} the
primary question and regards suspense as a crucial factor of
storytelling. Studies show that suspense is important for keeping readers'
attention \cite{khrypko11}, promotes readers' immersion and suspension
of disbelief \cite{hsu14}, and plays a big part in making stories
enjoyable and interesting \cite{oliver93, schraw01}. Computationally
less well understood, suspense has only sporadically been used in story
generation systems \cite{DBLP:conf/aaai/ONeillR14,
  cheong2014suspenser}.
  
Suspense, intuitively, is a feeling of anticipation that something
risky or dangerous will occur; this includes the idea both of
uncertainty and jeopardy. Take the play \textit{Romeo and Juliet:}
Dramatic suspense is created throughout --- the initial duel, the
meeting at the masquerade ball, the marriage, the
fight in which Tybalt is killed, and the sleeping potions leading to
the death of Romeo and Juliet. At each moment, the audience is
invested in something being at stake and wonders how it will end.

This paper aims to model suspense in computational terms, with the
ultimate goal of making it deployable in NLP systems that analyze or
generate narrative fiction. We start from the assumption that concepts
developed in psycholinguistics to model human language processing at
the word level \cite{Hale:01, hale2006uncertainty} can be generalised
to the story level to capture suspense, the Hale model. This assumption is supported
by the fact that economists have used similar concepts to model
suspense in games \citep{ely2015suspense,
  DBLP:conf/cogsci/LiBG18}, the Ely model. Common to both approaches is the idea that
suspense is a form of expectation: In games, we expect to win or lose instead
in stories, we expect that the narrative will end a certain way.

We will therefore compare two ways for modelling narrative suspense:
surprise, a backward-looking measure of how unexpected the current
state is given the story so far; and uncertainty reduction, a
forward-looking and measure of how unexpected the continuation of the
story is.  Both measures can be computed either directly over story
representations, or indirectly over the probability distributions over
such representations. We propose a hierarchical language model based
on Generative Pre-Training (GPT, \citealp{radford2018improving}) to
encode story-level representations and develop an inference scheme
that uses these representations to compute both surprise and
uncertainty reduction.  For evaluation, we use the WritingPrompt
corpus of short stories \cite{fan-etal-2018-hierarchical}, part of
which we annotate with human sentence-by-sentence judgements of
suspense. We find that surprise over representations and over probability distributions both predict suspense judgements. However uncertainty reduction over representations is better, resulting
in near human-level accuracy. We also show that our models can be used
to predict turning points, i.e., major narrative events, in movie
synopses \cite{papalampidi-etal-2019-movie}.

\section{Related Work}

% \paragraph{Plots and Structure} Most relevant work in the field
% takes a \emph{top-down} approach to plot structure, a recent example
% is on the turning points detection of films
% \citep{papalampidi-etal-2019-movie}, from theory
% \citet{freytag1896freytag} Pyramid is most known. Having specific
% stages or points suits a supervised approach to learning. There is
% however a alternative tradition, espoused influentially by
% \citet{rabkin1973narrative}, that puts \emph{narrative suspense} at
% the core of what makes a plot \footnote{See
% \citet{perreault2018universal} for a detailed literature
% review.}. \citep{boyd2009origin} framed the origin of storytelling
% as the battle for attention. This school is far more cognitive and
% based on low level scripts and plans \citet{schank1975scripts} than
% the \emph{top-down} plots
% structure. \citet{schmid2003narrativity,schmid2017eventfulness} puts
% low level \emph{eventfulness} with low level uncertainty and
% significant state change as the basis for narratology. The relevance
% is this approach is based on expectations of what will happen next
% is in principle similar to auto-regressive models learn, and so
% amenable in principle to learning and predicting with unsupervised
% methods without relying on expensive annotation.

%\paragraph{Paradox of Suspense} 

In narratology, uncertainty over outcomes is traditionally seen as
suspenseful \citep[e.g.,][]{o2013computational,
  zillmann1996psychology, abbott2008cambridge}. Other authors claim
that suspense can exist without uncertainty
\citep[e.g.,][]{smuts2008desire, hoeken2000suspense,
  gerrig1989suspense} and that readers feel suspense even when they
read a story for the second time \citep{delatorre2018confronting},
which is unexpected if suspense is uncertainty; this is referred to as
the paradox of suspense \citep{prieto1998paradox,
  yanal1996paradox}. Considering \textit{Romeo and Juliet} again, in
the first view suspense is motivated by primarily by uncertainty over
what will happen. Who will be hurt or killed in the fight? What will
happen after marriage?  However, at the beginning of the play we are
told \say{from forth the fatal loins of these two foes, a pair of
  star-crossed lovers take their life}, and so the suspense is more
about being invested in the plot than not knowing the outcome,
aligning more with the second view: suspense can exist without
uncertainty. We do not address the paradox of suspense directly in
this paper, but we are guided by the debate to operationalise methods
that encompass both views. The Hale model is closer to the traditional
model of suspense as being about uncertainty. In contrast, the Ely
model is more in line with the second view that uncertainty matters
less than consequentially different outcomes.

% For example the prologue of \emph{Romeo and
%   Juliet} summarise the entire plot ahead of the play. There are also
% issues of perspective between the character (Doxa) and reader view
% \citep{schmid2003narrativity}. To illustrate imagine a man with an axe
% bursts in a woman, this is surprising for both reader and
% characters. If the reader knows the axeman is about to burst in then
% this creates suspense as there is concern over what might happen to
% the woman. If the man bursts in and the kisses the woman, there is now
% suspense in the reader.

%\paragraph{Suspense in Generation} 

In NLP, suspense is studied most directly in natural language
generation, with systems such as Dramatis
\citep{DBLP:conf/aaai/ONeillR14} and Suspenser
\citep{cheong2014suspenser}, two planning-based story generators that
use the theory of \citet{gerrig1994readers} that suspense is created
when a protagonist faces obstacles that reduce successful
outcomes. Our approach, in contrast, models suspense using general
language models fine-tuned on stories, without planning and domain
knowledge. The advantage is that the model can be trained on large
volumes of available narrative text without requiring expensive
annotations, making it more generalisable.

% \paragraph{Synthetic Tasks} The \textit{ROC Cloze} understanding
% task \citep{mostafazadeh-etal-2016-corpus} attracted attention using
% neural methods, but had issues with cues undermining task difficulty
% \citep{cai2017pay,sharma2018tackling}. This illustrates the problem
% with synthetic datasets generally, but also these kinds of stories
% are anaemic. They lack, the real dramatic tension that is the basis
% of this paper, and so a natural corpus becomes a necessity.

%\paragraph{Characters and Plot} 

Other work emphasises the role of characters and their development in
story understanding \citep{bamman-etal-2014-bayesian,
  bamman2013learning, chaturvedi2017unsupervised, iyyer2016feuding} or
summarisation \citep{gorinski-lapata-2018-whats}. A further important
element of narrative structure is plot, i.e.,~the sequence of events
in which characters interact. Neural models have explicitly modelled
events \cite{martin2018event, harrison2017toward,
  rashkin-etal-2018-event2mind} or the results of actions
\citep{roemmele-gordon-2018-encoder, liu-etal-2018-demn,
  liu-etal-2018-narrative}. On the other hand, some neural generation
models \citep{fan-etal-2018-hierarchical} just use a hierarchical
model on top of a language model; our architecture follows this
approach.

%%%%%%%%%%%%%%%%%%%%%%%%%%%%%%%%%%%%%%%%%%%%%%%%%%%%%%%%%%%%%%%%%%%%%%%%%%%%%%%%
\section{Models of Suspense}
\label{sec:suspense}

\subsection{Definitions}
\label{sec:def}

In order to formalise measures of suspense, we assume that a story
consists of a sequence of sentences. These sentences are processed one
by one, and the sentence at the current timepoint $t$ is represented
by an embedding $e_t$ (see Section~\ref{sec:architecture} for how
embeddings are computed). Each embedding is associated with a
probability $P(e_t)$. Continuations of the story are represented by a
set of possible next sentences, whose embeddings are denoted
by~$e^i_{t+1}$.

The first measure of suspense we consider is surprise \cite{Hale:01},
which in the psycholinguistic literature has been successfully used to
predict word-based processing effort \citep{Demberg:Keller:08a,
  roark-etal-2009-deriving, DBLP:journals/corr/abs-1810-11481,
  DBLP:conf/cogsci/SchijndelL18}. Surprise is a backward-looking
predictor: it measures how unexpected the current word is given the
words that preceded it (i.e.,~the left context).  Hale formalises
surprise as the negative log of the conditional probability of the
current word. For stories, we compute surprise over sentences. As our
sentence embeddings $e_t$ include information about the left context
$e_1,\dots,e_{t-1}$, we can write \emph{Hale surprise} as:
\begin{equation}
\begin{aligned}
S^\text{Hale}_t = -\log P(e_t)
\label{eqn:con_surprise}
\end{aligned}
\end{equation}
An alternative measure for predicting word-by-word processing effort
used in psycholinguistics is entropy reduction
\cite{hale2006uncertainty}. This measure is forward-looking: it
captures how much the current word changes our expectations about the
words we will encounter next (i.e.,~the right context). Again, we
compute entropy at the story level, i.e., over sentences instead of
over words. Given a probability distribution over possible next
sentences $P(e^i_{t+1})$, we calculate the entropy of that
distribution. Entropy reduction is the change of that entropy from one
sentence to the next:
\begin{equation}
\begin{aligned}
%\begin{split}
H_t = - \sum_{i}{P(e^i_{t+1}) \log P(e^i_{t+1})} \\ 
U^\text{Hale}_t = H_{t-1} - H_t%\end{split}
\label{eqn:ent_surprise}
\end{aligned}
\end{equation}
Note that we follow \citet{frank2013uncertainty} in computing entropy
over surface strings, rather than over parse states as in Hale's
original formulation.

In the economics literature, \citet{ely2015suspense} have proposed two
measures that are closely related to Hale surprise and entropy
reduction. At the heart of their theory of suspense is the notion of
belief in an end state. Games are a good example: the state of a
tennis game changes with each point being played, making a win more or
less likely. Ely et al. define surprise as the amount of change from
the previous time step to the current time step. Intuitively, large
state changes (e.g.,~one player suddenly comes close to winning) are
more surprising than small ones. Representing the state at time $t$ as
$e_t$, \emph{Ely surprise} is defined as:
\begin{equation}
\begin{aligned}
S_t^\text{Ely} = (e_{t} - e_{t-1})^2
\label{eqn:ely_surp}
\end{aligned}
\end{equation}
Ely et al.'s approach can be adapted for modelling suspense in stories
if we assume that each sentence in a story changes the state (the
characters, places, events in a story,~etc.). States $e_t$ then become
sentence embeddings, rather than beliefs in end states, and Ely
surprise is the distance between the current embedding $e_t$ and the
previous embedding~$e_{t-1}$. In this paper, we will use L1 and L2
distances; other authors \cite{DBLP:conf/cogsci/LiBG18} experiment
with information gain and KL divergence, but found worse performance when
modelling suspense in games. Just like Hale surprise, Ely surprise
models backward-looking prediction, but over representations, rather
than over probabilities.

Ely et al. also introduce a measure of forward-looking prediction,
which they define as the expected difference between the current
state $e_t$ and the next state~$e_{t+1}$:
\begin{equation}
\begin{aligned}
U^\text{Ely}_t = \mathop{\mathbb{E}}[(e_t - e^i_{t+1})^2] \\
=  \sum_{i} P(e^i_{t+1})(e_t - e^i_{t+1})^2
\label{eqn:ely_susp}
\end{aligned}
\end{equation}
This is closely related to Hale entropy reduction, but again the
entropy is computed over states (sentence embeddings in our case),
rather than over probability distributions. Intuitively, this measure
captures how much the uncertainty about the rest of the story is
reduced by the current sentence. We refer to the forward-looking
measures in Equations~(\ref{eqn:ent_surprise})
and~(\ref{eqn:ely_susp}) as \emph{Hale and Ely uncertainty reduction,}
respectively.

Ely et al. also suggest versions of their measures in which each state
is weighted by a value $\alpha_t$, thus accounting for the fact that
some states may be more inherently suspenseful than others:
\begin{equation}
\begin{aligned}
S_{t}^{\alpha\text{Ely}} = \alpha_t(e_{t} - e_{t-1})^2\\
\label{eqn:ely_surp_ext}
U^{\alpha\text{Ely}}_t = \mathop{\mathbb{E}}[\alpha_{t+1}(e_t - e^i_{t+1})^2]
\end{aligned}
\end{equation}
We stipulate that sentences with high emotional valence are more
suspenseful, as emotional involvement heightens readers' experience of
suspense. This can be captured in Ely et al.'s framework by assigning
the $\alpha$s the scores of a sentiment classifier.

% Figure~\ref{fig:suspense_illustration} provides an illustration of
% the difference between suspense and uncertainty reduction, and between
% measures computed over probabilities and over representations.

% \begin{figure}[tb]
% \centering
% \includegraphics[width=0.35\textwidth]{images/Suspense_2.pdf}
% \caption{Illustration of suspense and surprise, $x$ is time, $y$
%   distance is how much states differ, circle size is the probability
%   of the path occurring; \textit{green} is the actual path of the
%   story, and \textit{red} alternative outcomes.}
% \label{fig:suspense_illustration}
% \end{figure}

\subsection{Modelling Approach}
\label{sec:model_approach}

We now need to show how to compute the surprise and uncertainty
reduction measures introduced in the previous section. This involves
building a model that processes stories sentence by sentence, and
assigns each sentence an embedding that encodes the sentence and its
preceding context, as well as a probability. These outputs can then be
used to compute a surprise value for the sentence.

Furthermore, the model needs to be able to generate a set of possible
next sentences (story continuations), each with an embedding and a
probability. Generating upcoming sentences is potentially very
computationally expensive since the number of continuations grows
exponentially with the number of future time steps. As an
alternative, we can therefore sample possible next sentences from a
corpus and use the model to assign them embeddings and
probabilities. Both of these approaches will produce sets of upcoming
sentences, which we can then use to compute uncertainty
reduction. While we have so far only talked about the next sentences,
we will also experiment with uncertainty reduction computed using
longer rollouts.

% Embeddings implicitly encode information about the situation
% \citep{DBLP:conf/acl/BaroniBLKC18,DBLP:conf/nips/LevyG14}. The
% advantages: It is theoretically guided by literary and cognitive
% theory. Success of language models such as ELMO
% \citep{peters-etal-2018-deep} and BERT \citep{devlin-etal-2019-bert}
% have shown that highly powerful models can be trained
% auto-regressively eliminates much of the bottleneck on available
% training data, and also lends itself to cheaply to broader application
% in a way a supervised model does not.

%%%%%%%%%%%%%%%%%%%%%%%%%%%%%%%%%%%%%%%%%%%%%%%%%%%%%%%%%%%%%%%%%%%%%%%%%%%%%%%%
\section{Model}
\label{sec:architecture}

\subsection{Architecture}

\begin{figure}[tb]
\includegraphics[trim={0.82cm 0cm 0.5cm 0cm},clip,width=0.5\textwidth]{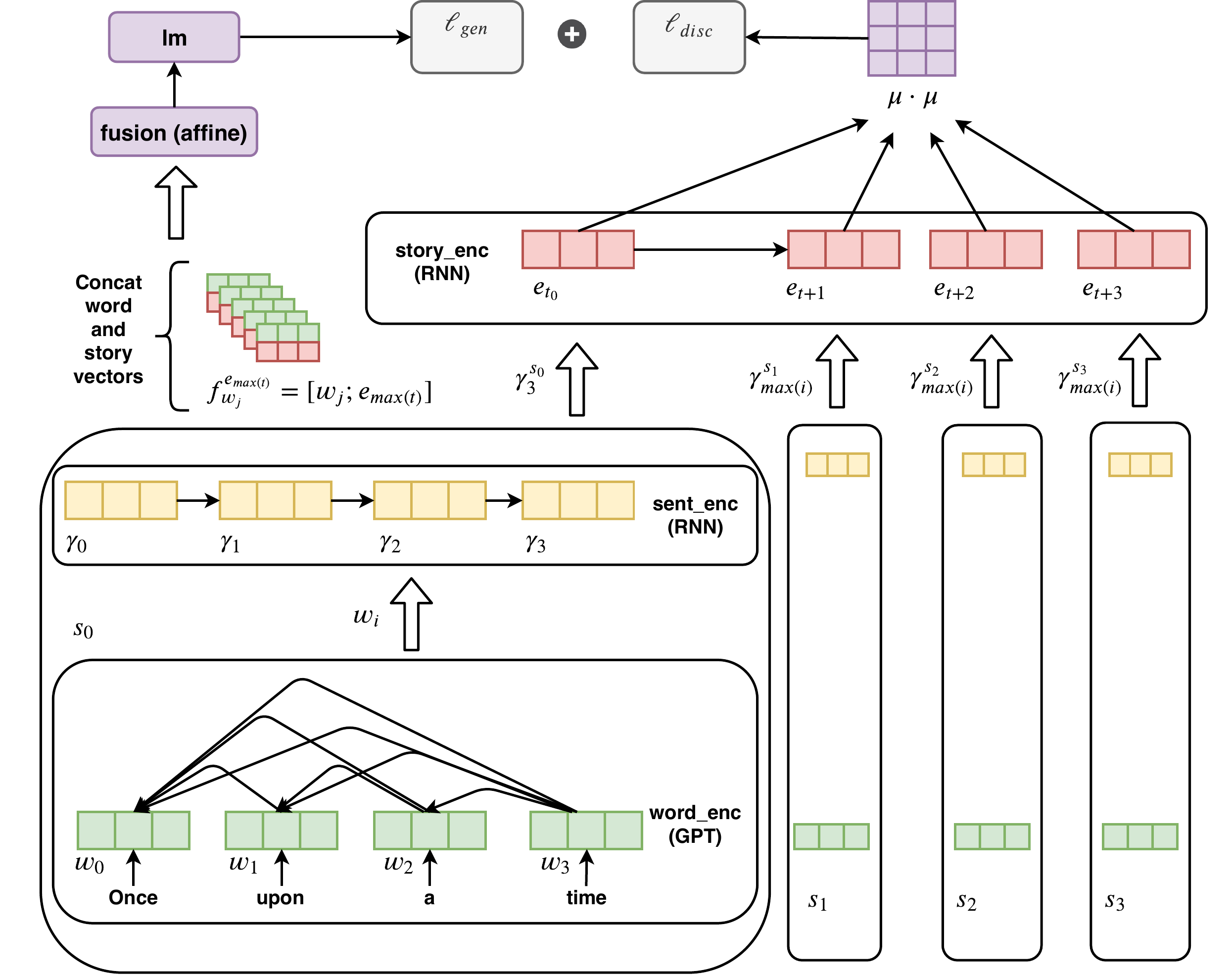}
\vspace*{-3ex}
\caption{Architecture of our hierarchical model. See text for
  explanation of the components \emph{word\_enc,} \emph{sent\_enc,}
  and \emph{story\_enc}.}
\label{fig:suspense_arch}
\end{figure}

Our overall approach leverages contextualised language models, which
are a powerful tool in NLP when pretrained on large amounts of text
and fine tuned on a specific task \citep{peters-etal-2018-deep,
  devlin-etal-2019-bert}. Specifically, we use Generative Pre-Training
(GPT, \citealp{radford2018improving}), a model which has proved
successful in generation tasks \citep{radford2019language,
  see2019massively}.

\paragraph{Hierarchical Model} 

Previous work found that hierarchical models show strong performance
in story generation \citep{fan-etal-2018-hierarchical} and
understanding tasks \citep{cai2017pay}. The language model and
hierarchical encoders we use are uni-directional, which matches the
incremental way in which human readers process stories when they
experience suspense.

Figure~\ref{fig:suspense_arch} depicts the architecture of our
hierarchical model.\footnote{Model code and scripts for evaluation are
  available at
  \url{https://github.com/dwlmt/Story-Untangling/tree/acl-2020-dec-submission}}
It builds a chain of representations that anticipates what will come
next in a story, allowing us to infer measures of suspense. For a
given sentence, we use GPT as our word encoder (\emph{word\_enc} in
Figure~\ref{fig:suspense_arch}) which turns each word in a sentence
into a word embedding~$w_i$. Then, we use an RNN (\emph{sent\_enc}) to
turn the word embeddings of the sentences into a sentence
embedding~$\gamma_i$. Each sentence is represented by the hidden state
of its last word, which is then fed into a second RNN
(\emph{story\_enc}) that computes a story embedding. The overall story
representation is the hidden state of its last sentence. Crucially,
this model also gives us $e_t$, a contextualised representation of the
current sentence at point~$t$ in the story, to compute surprise and
uncertainty reduction.

Model training includes a generative loss $\ell_{gen}$ to improve the
quality of the sentences generated by the model. We concatenate the
word representations $w_j$ for all word embeddings in the latest
sentence with the latest story embedding $e_{\max(t)}$. This is run
through affine ELU layers to produce enriched word embedding
representations, analogous to the Deep Fusion model
\citep{DBLP:journals/corr/GulcehreFXCBLBS15}, with story state instead
of a translation model. The related Cold Fusion approach
\citep{DBLP:conf/interspeech/SriramJSC18} proved inferior.

\paragraph{Loss Functions} 

To obtain the discriminatory loss $\ell_{disc}$ for a particular
sentence~$s$ in a batch, we compute the dot product
%
%\footnote{Cosine was also tried but performed worse in training.}
%
of all the story embeddings $e$ in the batch, and then take the
cross-entropy across the batch with the correct next sentence:
\begin{equation}
\begin{aligned}
\ell_{disc}(e_{t+1}^{i=s}) = -\log \frac{\exp(e_{t+1}^{i=s} \cdot e_{t})}{\sum_i \exp(e_{t+1}^{i}  \cdot e_{t})}
\label{eqn:quickthoughts_loss}
\end{aligned}
\end{equation}
Modelled on Quick Thoughts \citep{logeswaran2018an}, this forces the
model to maximise the dot product of the correct next sentence versus
other sentences in the same story, and negative examples from other
stories, and so encourages representations that anticipate what
happens next.

The generative loss in Equation~(\ref{eqn:gen_loss}) is a standard LM
loss, where $w_j$ is the GPT word embeddings from the sentence and
$e_{\max(t)}$ is the story context that each word is concatenated
with:
\begin{equation}
%\begin{aligned}
\ell_{gen} = - \sum_{j} \log P({w_j | w_{j-1}, w_{j-2}, \dots; e_{\max(t)})}
\label{eqn:gen_loss}
%\end{aligned}
\end{equation}
The overall loss is $\ell_{disc} + \ell_{gen}$. More advanced
generation losses \citep[e.g.,][]{DBLP:journals/corr/abs-1905-12616}
could be used, but are an order of magnitude slower.

\subsection{Inference} 

We compute the measures of surprise and uncertainty reduction
introduced in Section~\ref{sec:def} using the output of the story
encoder \emph{story\_enc}. In addition to the contextualised sentence
embeddings $e_t$, this requires their probabilities $P(e_t)$, and a
distribution over alternative continuations~$P(e^i_{t+1})$.

We implement a recursive beam search over a tree of future sentences
in the story, looking between one and three sentences ahead
(rollout). The probability is calculated using the same method as the
discriminatory loss, but with the cosine similarity rather than the dot product
of the embeddings $e_t$ and $e^i_{t+1}$ fed into a softmax
function. We found that cosine outperformed dot product on inference
as the resulting probability distribution over continuations is less
concentrated.

%%%%%%%%%%%%%%%%%%%%%%%%%%%%%%%%%%%%%%%%%%%%%%%%%%%%%%%%%%%%%%%%%%%%%%%%%%%%%%%%
\section{Methods}
\label{sec:methods}

\paragraph{Dataset}
\label{sec:datasets}

The overall goal of this work is to test whether the psycholinguistic
and economic theories introduced in Section~\ref{sec:suspense} are
able to capture human intuition of suspense. For this, it is important
to use actual stories which were written by authors with the aim of
being engaging and interesting. Some of the story datasets used in NLP
do not meet this criterion; for example ROC Cloze
\citep{mostafazadeh-etal-2016-corpus} is not suitable because the
stories are very short (five sentences), lack naturalness, and are
written by crowdworkers to fulfill narrow objectives, rather than to
elicit reader engagement and interest. A number of authors have also
pointed out technical issues with such artificial corpora
\citep{cai2017pay, sharma2018tackling}.

Instead, we use WritingPrompts \cite{fan-etal-2018-hierarchical}, a
corpus of circa 300k short stories from the \texttt{/r/WritingPrompts}
subreddit. These stories were created as an exercise in creative
writing, resulting in stories that are interesting, natural, and of
suitable length. The original split of the data into 90\% train, 5\%
development, and 5\% test was used. Pre-processing steps are described
in Appendix~\ref{app:a}.

\paragraph{Annotation} 

To evaluate the predictions of our model, we selected $100$ stories
each from the development and test sets of the WritingPrompts corpus,
such that each story was between $25$ and $75$ sentence in
length. Each sentence of these stories was judged for narrative
suspense; five master workers from Amazon Mechanical Turk annotated
each story after reading instructions and completing a training phase.
They read one sentence at a time and provided a suspense judgement
using the five-point scale consisting of Big Decrease in suspense (1\%
of the cases), Decrease (11\%), Same (50\%), Increase (31\%), and Big
Increase (7\%). In contrast to prior work
\citep{delatorre2018confronting}, a relative rather than absolute
scale was used. Relative judgements are easier to make while reading,
though in practice, the suspense curves generated are very similar,
with a long upward trajectory and flattening or dip near the end.
After finishing a story, annotators had to write a short summary of
the story.

In the instructions, suspense was framed as \emph{dramatic tension},
as pilot annotations showed that the term suspense was too closely
associated with murder mystery and related genres. Annotators were
asked to take the character's perspective when reading to achieve
stronger inter-annotator agreement and align closely with literary
notions of suspense. During training, all workers had to annotate a
test story and achieve 85\% accuracy before they could continue.  Full
instructions and the training story are in Appendix~\ref{app:b}.

The inter-annotator agreement $\alpha$
\citep{krippendorff2011computing} was $0.52$ and $0.57$ for the
development and test sets, respectively. Given the inherently
subjective nature of the task, this is substantial agreement. This was
achieved after screening out and replacing annotators who had low
agreement for the stories they annotated (mean $\alpha < 0.35$),
showed suspiciously low reading times (mean RT $<600$~ms per
sentence), or whose story summaries indicated low-quality annotation.

\begin{table}[tb]
\centering
\begin{tabular}{@{}lllll@{}}
\toprule
                   & \textbf{GRU} & \textbf{LSTM}  \\ \midrule
Loss      & 5.84       & 5.90      \\
Discriminatory  Accuracy  & 0.55       & 0.54        \\
Discriminatory Accuracy $k=10$ & 0.68       & 0.68         \\
Generative Accuracy    & 0.37       & 0.46        \\
Generative Accuracy $k=10$  & 0.85       & 0.85       \\
Cosine Similarity       & 0.48       & 0.50       \\
L2 Distance        & 1.73       & 1.59    \\
Number of Epochs    & 4          & 2      \\ \bottomrule
\end{tabular}
\caption{For accuracy the baseline probability is 1 in 99; $k=10$ is
  the accuracy of the top 10 sentences of the batch. From the best
  epoch of training on the WritingPrompts development set.}
\label{tab:tech_results}
\end{table}

\paragraph{Training and Inference}

The training used SGD with Nesterov momentum
\citep{sutskever2013importance} with a learning rate of $0.01$ and a
momentum of $0.9$. Models were run with early stopping based on the
mean of the accuracies of training tasks. For each batch, $50$ sentence blocks from two
different stories were chosen to ensure that the negative examples in
the discriminatory loss include easy (other stories) and difficult
(same story) sentences.

We used the pretrained GPT weights but fine-tuned the encoder and
decoder weights on our task. For the RNN components of our
hierarchical model, we experimented with both GRU
\citep{chung2015gated} and LSTM
\citep{DBLP:journals/neco/HochreiterS97} variants. The GRU model had
two layers in both \emph{sen\_enc} and \emph{story\_enc}; the LSTM
model had four layers each in \emph{sen\_enc} and
\emph{story\_enc}. Both had two fusion layers and the size of the
hidden layers for both model variants was $768$. We give the results
of both variants on the tasks of sentence generation and sentence
discrimination in Table~\ref{tab:tech_results}. Both perform
similarly, with slightly worse loss for the LSTM variant, but faster
training and better generation accuracy. Overall, model performance is
strong: the LSTM variant picks out the correct sentence 54\% of the
time and generates it 46\% of the time. This indicates that our
architecture successfully captures the structure of stories.

At inference time, we obtained a set of story continuations either by
random sampling or by generation. Random sampling means that $n$
sentences were selected from the corpus and used as continuations. For
generation, sentences were generated using top-$k$ sampling (with
$k=50$) using the GPT language model and the approach of
\citet{radford2019language}, which generates better output than beam
search \citep{holtzman-etal-2018-learning} and can outperform a
decoder \citep{see2019massively}. For generation, we used up to $300$
words as context, enriched with the story sentence embeddings from the
corresponding points in the story. For rollouts of one sentence, we
generated $100$ possibilities at each step; for rollouts of two, $50$
possibilities and rollouts of three, $25$ possibilities. This keeps what
is an expensive inference process manageable.

\paragraph{Importance} 

We follow Ely et al. in evaluating weighted versions of their surprise
and uncertainty reduction measure $S_t^{\alpha\text{Ely}}$ and
$U_t^{\alpha\text{Ely}}$ (see Equation~(\ref{eqn:ely_surp_ext})). We
obtain the $\alpha_t$ values by taking the sentiment scores assigned
by the VADER sentiment classifier \citep{hutto2014vader} to each
sentence and multiplying them by $1.0$ for positive sentiment and
$2.0$ for negative sentiment. The stronger negative weighting reflects
the observation that negative consequences can be more important than
positive ones \citep{o2013computational, kahneman2013prospect}.

\paragraph{Baselines} 

We test a number of baselines as alternatives to surprise and
uncertainty reduction derived from our hierarchical model. These
baselines also reflect how much change occurs from one sentence to the
next in a story: WordOverlap is the Jaccard similarity between the two
sentences, GloveSim is the cosine similarity between the averaged
Glove \citep{pennington2014glove} word embeddings of the two
sentences, and GPTSim is the cosine similarity between the GPT
embeddings of the two sentences. The $\alpha$ baseline is the weighted
VADER sentiment score.

%%%%%%%%%%%%%%%%%%%%%%%%%%%%%%%%%%%%%%%%%%%%%%%%%%%%%%%%%%%%%%%%%%%%%%%%%%%%%%%%
\section{Results}
\label{sec:results}

\subsection{Narrative Suspense}

\paragraph{Task} 

The annotator judgements are relative (amount of decrease/increase in
suspense from sentence to sentence), but the model predictions are
absolute values.  We could convert the model predictions
into discrete categories, but this would fail to capture the overall
arc of the story. Instead, we convert the relative judgements into
absolute suspense values, where $J_t = j_1 + \dots + j_t$ is the
absolute value for sentence $t$ and $j_1, \dots, j_t$ are the relative
judgements for sentences~1 to $t$.  We use $-0.2$ for Big Decrease,
$-0.1$ for Decrease, $0$ for Same, $0.1$ for Increase, and $0.2$ for
Big Increase.\footnote{These values were fitted with predictions (or
  cross-worker annotation) using 5-fold cross validation and an L1
  loss to optimise the mapping. A constraint is placed so that Same is
  $0$, increases are positive and decreases are negative with a
  minimum $0.05$ distance between.} Both the absolute suspense
judgements and the model predictions are normalised by converting them
to $z$-scores.

To compare model predictions and absolute suspense values, we use
Spearman's $\rho$ \citep{sen1968estimates} and Kendall's $\tau$
\citep{kendall1975rank}. Rank correlation is preferred because we are
interested in whether human annotators and models view the same part
of the story as more or less suspenseful; also, rank correlation
methods are good at detecting trends.
%
%L1 or L2 distance based measures perform badly on co-integration
%\citep{murray1994drunk} methods such as Engle-Granger
%\citep{10.2307/1913236} are unstable.
%
We compute $\rho$ and $\tau$ between the model predictions and the
judgements of each of the annotators (i.e.,~five times for five
annotators), and then take the average. We then average these values
again over the 100~stories in the test or development sets. As the
human upper bound, we compute the mean pairwise correlation of the
five annotators.

\begin{figure}[tb]
\includegraphics[trim={1.5cm 0.5cm 2.5cm 2.5cm},clip,width=0.50\textwidth]{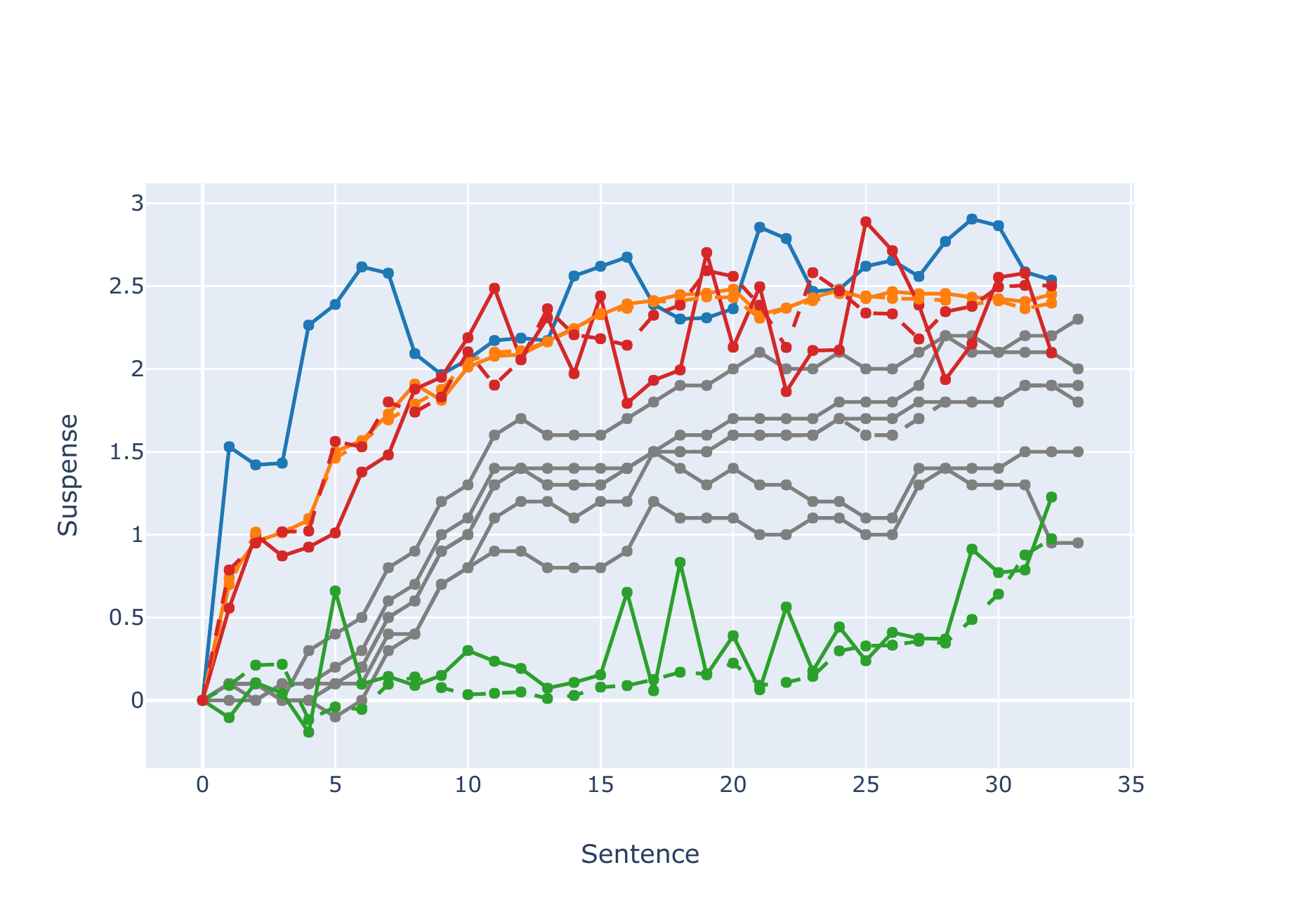}
\vspace*{-4ex}
\caption{Story 27,
\textbf{\textcolor{annotatedcolor}{Human}},
\textbf{\textcolor{surpriseentropycolor}{$S^{\text{Hale}}$}},
\textbf{\textcolor{surprisecolor}{$S^{\text{Ely}}$}},
\textbf{\textcolor{suspensecolor}{$U^{\text{Ely}}$}},
\textbf{\textcolor{suspensestatecolor}{$U^{\alpha\text{Ely}}$.}} Solid
lines: generated alternative continuations, dashed lines: sampled
alternative continuations.}
\label{fig:wp_27_main}
\end{figure}

\begin{table}[ht!]
\centering
\begin{tabular}{@{}ll@{}ccc@{}}
\toprule
\textbf{Prediction}  & \textbf{Model} & \textbf{Rollout} & \textbf{$\tau$ $\uparrow$}           & \textbf{$\rho$ $\uparrow$}           \\ \midrule
Human                &                &               & \textbf{.553}                        & \textbf{.614}                        \\ \midrule
Baselines            & WordOverlap    & 1             & .017                                 & .026                                 \\
                     & GloveSim       & 1             & .017                                 & .029                                 \\
                     & GPTSim         & 1             & .021                                 & .031                                 \\
                     & $\alpha$       & 1             & \textbf{.024}                        & \textbf{.036}                        \\ \midrule
$S^\text{Hale}$-Gen   & GRU            & 1             & .145                                 & .182                                 \\
                     & LSTM           & 1             & \textbf{.434}                        & \textbf{.529}                        \\ \midrule
$S^\text{Hale}$-Cor   & GRU            & 1             & .177                                 & .214                                 \\
                     & LSTM           & 1             & \textbf{.580}                        & \textbf{.675}                        \\ \midrule
$U^\text{Hale}$-Gen   & GRU            & 1             & \textbf{.036}                        & \textbf{.055}                        \\
                     & LSTM           & 1             & .009                                 & .016                                 \\ \midrule
$U^\text{Hale}$-Cor   & GRU            & 1             & .048                                 & .050                                 \\
                     & LSTM           & 1             & \textbf{.066}                        & \textbf{.094}                        \\ \midrule
$S^{\text{Ely}}$        & GRU            & 1             & \textbf{.484}                        & \textbf{.607}                        \\
                     & LSTM           & 1             & .427                                 & .539                                 \\ \midrule
$S^{\alpha\text{Ely}}$   & GRU            & 1             & .089                                 & .123                                 \\
                     & LSTM           & 1             & \textbf{.115}                        & \textbf{.156}                        \\ \midrule
$U^{\text{Ely}}$-Gen   & GRU             & 1             & .241                                 & .161                                 \\
                     &                & 2             & .304                                 & .399                                 \\
                     & LSTM           & 1             & {\color[HTML]{9A0000} \textbf{.610}} & {\color[HTML]{9A0000} \textbf{.698}} \\
                     &                & 2             & .393                                 & .494                                 \\ \midrule
$U^{\text{Ely}}$-Cor    & GRU            & 1             & .229                                 & .264                                 \\
                     &                & 2             & .512                                 & .625                                 \\
                     &                & 3             & .515                                 & .606                                 \\
                     & LSTM           & 1             & \textbf{.594}                        & \textbf{.678}                        \\
                     &                & 2             & .564                                 & .651                                 \\
                     &                & 3             & .555                                 & .645                                 \\ \midrule
$U^{\alpha\text{Ely}}$-Gen & GRU          & 1             & .216                                 & .124                                 \\
                     &                & 2             & .219                                 & .216                                 \\
                     & LSTM           & 1             & \textbf{.474}                        & \textbf{.604}                        \\
                     &                & 2             & .316                                 & .418                                 \\ \midrule
$U^{\alpha\text{Ely}}$-Cor & GRU          & 1             & .205                                 & .254                                 \\
                     &                & 2             & .365                                 & .470                                 \\
                     & LSTM           & 1             & \textbf{.535}                        & \textbf{.642}                        \\
                     &                & 2             & .425                                 & .534                                 \\ \bottomrule
\end{tabular}
\caption{Development set results for WritingPrompts for generated
  (Gen) or corpus sampled (Cor) alternative continuations; $\alpha$
  indicates sentiment weighting. \textbf{Bold:} best model in a given category;
  {\color[HTML]{9A0000}\textbf{red:}} best model overall.} 
\label{tab:dev_wp_res}
\end{table}

\begin{table}[tb]
\centering
\begin{tabular}{@{}lrr@{}}
\toprule
\textbf{Prediction}        & \multicolumn{1}{c}{$\tau$ $\uparrow$}                   & \multicolumn{1}{c}{$\rho$ $\uparrow$}                     \\ \midrule
Human                      & {\color[HTML]{9A0000} \textbf{.652}} (.039)     & {\color[HTML]{9A0000} \textbf{.711}} (.033)     \\ \midrule
$S^\text{Hale}$-Gen               & .407                                 (.089)     & .495                                 (.081)     \\
$S^\text{Hale}$-Cor               & \textbf{.454}                        (.085)     & \textbf{.523}                        (.079)     \\
$U^\text{Hale}$-Gen               & .036                                 (.102)     & .051                                 (.102)     \\
$U^\text{Hale}$-Cor               & .061                                 (.100)     & .088                                 (.101)     \\ \midrule
$S^{\text{Ely}}$              & .391                                 (.092)     & .504                                 (.082)     \\
%$S^{\alpha\text{Ely}}$     & .090                                 (.102)     & .130                                 (.102)     \\
$U^{\text{Ely}}$-Gen          & \textbf{.620}                         (.067)     & \textbf{.710}                        (.053)     \\
$U^{\text{Ely}}$-Cor          & .605                                  (.069)     & .693                                 (.056)     \\
%$U^{\alpha\text{Ely}}$-Gen & .450                                 (.085)     & .580                                 (.072)     \\
%$U^{\alpha\text{Ely}}$-Cor & .538                                 (.077)     & .646                                 (.064)     \\
\bottomrule
\end{tabular}
\caption{Test set results for WritingPrompts for generated
  (Gen) or corpus sampled (Cor) continuations. LSTM with rollout 
  one; brackets: confidence intervals.}
\label{tab:test_wp_res}
\end{table}

\paragraph{Results} 

Figure~\ref{fig:wp_27_main} shows surprise and uncertainty reduction
measures and human suspense judgements for an example story (text and
further examples in Appendix~\ref{app:c}).
% \footnote{The plots all have an initial jump which likely reflects
% the lack of model context but doesn't affect the rank correlation
% much.} 
We performed model selection using the correlations on the development
set, which are given in Table~\ref{tab:dev_wp_res}. We experimented
with all the measures introduced in Section~\ref{sec:def}, computing
sets of alternative sentences either using generated continuations
(Gen) or continuations sampled from the corpus (Cor), except for
$S^\text{Ely}$, which can be computed without alternatives. We
compared the LSTM and GRU variants (see
Section~\ref{sec:architecture}) and experimented with rollouts of up
to three sentences. We tried L1 and L2 distance for the Ely measures,
but only report L1, which always performed better.

\paragraph{Discussion}

On the development set (see Table~\ref{tab:dev_wp_res}), we observe
that all baselines perform poorly, indicating that distance between
simple sentence representations or raw sentiment values do not model
suspense. We find that Hale surprise $S^\text{Hale}$ performs well,
reaching a maximum $\rho$ of .675 on the development set.  Hale
uncertainty reduction $U^\text{Hale}$, however, performs consistently
poorly. Ely surprise $S^\text{Ely}$ also performs well, reaching as
similar value as Hale surprise. Overall, Ely uncertainty reduction
$U^\text{Ely}$ is the strongest performer, with $\rho = .698$,
numerically outperforming the human upper bound.

Some other trends are clear from the development set: using GRUs
reduces performance in all cases but one; rollout of more than one
never leads to an improvement; sentiment weighting (prefix $\alpha$ in
the table) always reduces performance, as it introduces considerable
noise (see Figure~\ref{fig:wp_27_main}). We therefore eliminate the
models that correspond to these settings when we evaluate on the test
set.

For the test set results in Table~\ref{tab:test_wp_res} we also report
upper and lower confidence bounds computed using the Fisher
$Z$-transformation ($p < 0.05$).  On the test set, $U^{\text{Ely}}$
again is the best measure, with a correlation statistically
indistinguishable from human performance (based on CIs).  We find that
absolute correlations are higher on the test set, presumably
reflecting the higher human upper bound.

Overall, we conclude that our hierarchical architecture successfully
models human suspense judgements on the WritingPrompts dataset. The
overall best predictor is $U^{\text{Ely}}$, uncertainty reduction
computed over story representations. This measure combines the
probability of continuation ($S^\text{Hale}$) with distance between
story embeddings ($S^{\text{Ely}}$), which are both good predictors in
their own right. This finding supports the theoretical claim that
suspense is an expectation over the change in future states of a game
or a story, as advanced by \citet{ely2015suspense}.

\subsection{Movie Turning Points}

\paragraph{Task and Dataset}

An interesting question is whether the peaks in suspense in a story
correspond to important narrative events. Such events are sometimes
called turning points (TPs) and occur at certain positions in a movie
according to screenwriting theory \cite{cutting2016narrative}. A corpus
of movie synopses annotated with turning points is available in the
form of the TRIPOD dataset \citep{papalampidi-etal-2019-movie}. We can
therefore test if surprise or uncertainty reduction predict TPs in
TRIPOD. As our model is trained on a corpus of short stories, this
will also serve as an out-of-domain evaluation.

% \paragraph{Dataset} 
% 
% Since the model is unsupervised, one interesting point of evaluation
% is if suspense predictions align a plan-driven structure. This is
% evaluated using Turning Points on Movie synopsis using the TRIPOD
% dataset \citep{papalampidi-etal-2019-movie}. Synopses are quite a
% different medium from short stories, and so it is also a test of how
% the model in a different domain. Models finetuned using the CMU movie
% corpus \citep{bamman2013learning} of circa 42K movies but produced
% slightly worse results.

%
\begin{table}[tb]
\centering
\begin{tabular}{@{}lrr@{}}
\toprule
            & \multicolumn{1}{c}{\textbf{Dev $D$} $\downarrow$}            & \multicolumn{1}{c}{\textbf{Test $D$} $\downarrow$}                    \\ \midrule
Human       & Not reported          & \color[HTML]{9A0000}{\textbf{4.30 (3.43)}}         \\ \midrule
%Random        & NA           & 37.8 (25.3)        \\
Theory Baseline & 9.65 (0.94)          & 7.47 (3.42)          \\
TAM         & \color[HTML]{9A0000}{\textbf{7.11 (1.71)}} & 6.80 (2.63) \\ \midrule
WordOverlap          & 13.9 (1.45)          & 12.7 (3.13)         \\
GloveSim          & \textbf{10.2 (0.74)}          & \textbf{10.4 (2.54)}         \\ 
GPTSim          & 16.8 (1.47)       & 18.1 (4.71)         \\ 
$\alpha$          & 11.3 (1.24)         & 11.2 (2.67)         \\  \midrule
$S^\text{Hale}$-Gen        & \textbf{8.27 (0.68)}          & \textbf{8.72 (2.27)}          \\
$U^\text{Hale}$-Gen        & 10.9 (1.02)          & 10.69 (3.66)          \\
\midrule
$S^{\text{Ely}}$    & 9.54 (0.56)          & 9.01 (1.92)          \\
$S^{\alpha\text{Ely}}$       & 9.95 (0.78)          & 9.54 (2.76)          \\
$U^{\text{Ely}}$-Gen       & 8.75 (0.76)          & 8.38 (1.53)          \\
$U^{\text{Ely}}$-Cor       & 8.74 (0.76)         & 8.50 (1.69)          \\
$U^{\alpha\text{Ely}}$-Gen   & 8.80 (0.61) & 7.84 (3.34) \\
$U^{\alpha\text{Ely}}$-Cor    &\textbf{8.61 (0.68)} & \textbf{7.78 (1.61)} \\ \bottomrule
\end{tabular}
\caption{TP prediction on the TRIPOD
  development and test sets. $D$ is the normalised distance to the
  gold standard; CI in brackets.}
\label{tab:turn_point_res}
\end{table}

\begin{figure}[tb]
\includegraphics[trim={1.5cm 0.5cm 2.5cm 2.5cm},clip,width=0.50\textwidth]{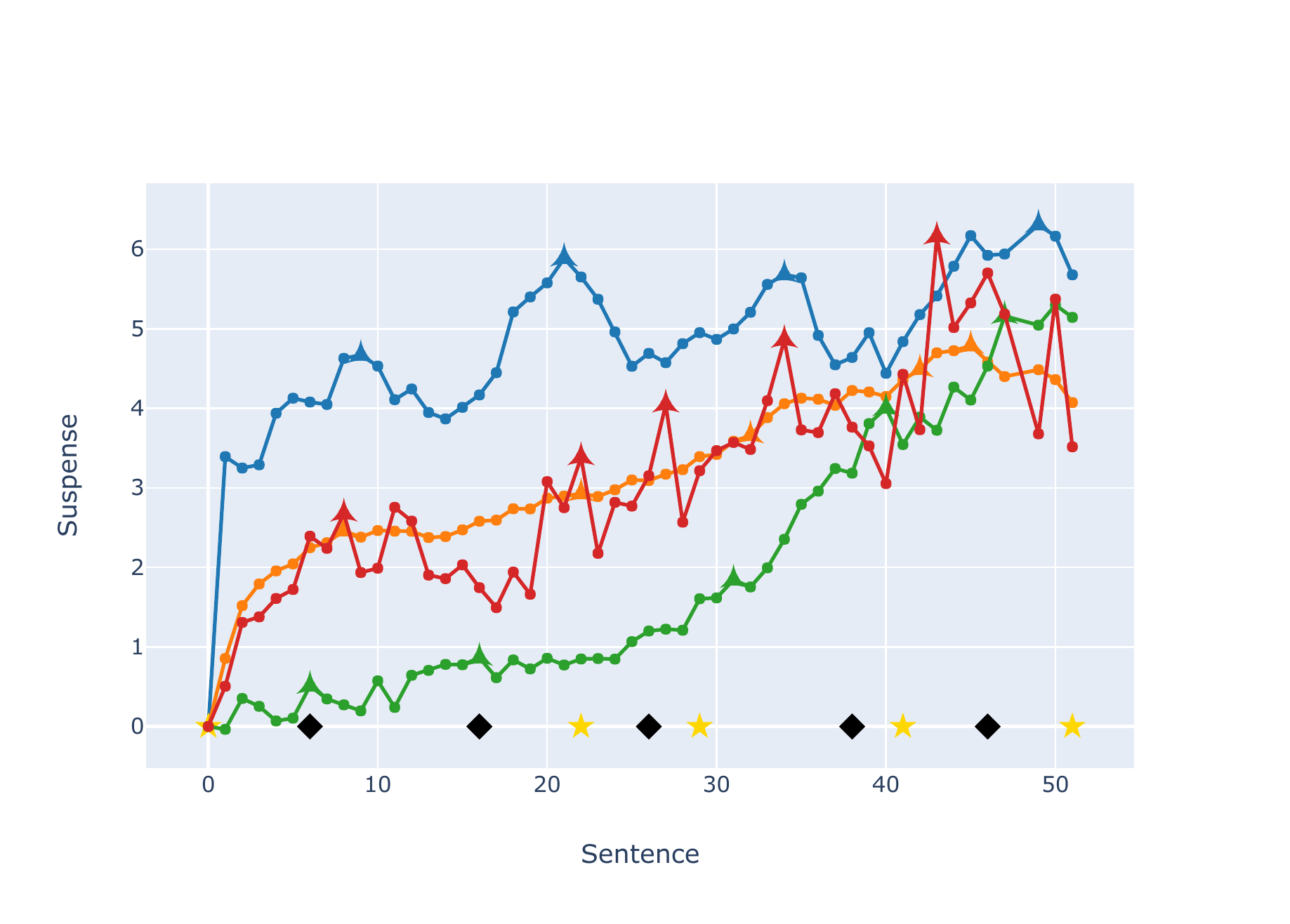}
\vspace*{-4ex}
\caption{Movie \href{https://www.imdb.com/title/tt0179626/}{15
    Minutes},
  \textbf{\textcolor{surpriseentropycolor}{$S^{\text{Hale}}$}},
  \textbf{\textcolor{surprisecolor}{$S^{\text{Ely}}$}},
  \textbf{\textcolor{suspensecolor}{$U^{\text{Ely}}$}},
  \textbf{\textcolor{suspensestatecolor}{$U^{\alpha\text{Ely}}$}},
  $\medblackdiamond$ theory baseline, $ \color{yellow} \medstar$~TP annotations, triangles are
  predicted TPs.}
\label{fig:tp_test_8_main}
\end{figure}

\citet{papalampidi-etal-2019-movie} assume five TPs: 1.~Opportunity,
2.~Change of Plans, 3.~Point of no Return, 4.~Major Setback, and
5.~Climax. They derive a prior distribution of TP positions from their
test set, and use this to constrain predicted turning points to
windows around these prior positions. We follow this approach and
select as the predicted TP the sentence with the highest surprise or
uncertainty reduction value within a given constrained window. We
report the same baselines as in the previous experiment, as well as
the Theory Baseline, which uses screenwriting theory to predict where
in a movie a given TP should occur (e.g.,~Point of No Return
theoretically occurs 50\% through the movie). This baseline is hard to
beat \cite{papalampidi-etal-2019-movie}.

% It appears to reveal some structural relations where suspense peaks
% for TPs 1 and 2 occur consistently just after the TPs, and for TPs
% 3, 4, and 5 before; the structural relations between suspense and
% plot structure is a promising avenue for future work.

\paragraph{Results and Discussion} 

Figure~\ref{fig:tp_test_8_main} plots both gold standard and predicted
TPs for a sample movie synopsis (text and further examples in
Appendix~\ref{app:d}). The results on the TRIPOD development and test
sets are reported in Table~\ref{tab:turn_point_res} (we report both
due to the small number of synopses in TRIPOD). We use our best LSTM
model with a of rollout of one; the distance measure for Ely surprise
and uncertainty reduction is now L2 distance, as it outperformed L1 on
TRIPOD. We report results in terms of~$D$, the normalised distance
between gold standard and predicted TP~positions.

On the test set, the best performing model with $D = 7.78$ is
$U^{\alpha\text{Ely}}$-Cor, with $U^{\alpha\text{Ely}}$-Gen only
slightly worse. It is outperformed by TAM, the best model of
\citet{papalampidi-etal-2019-movie}, which however requires TP
annotation at training time.  $U^{\alpha\text{Ely}}$-Cor is close to
the Theory Baseline on the test set, an impressive result given that our model has no TP supervision
and is trained on a different domain. The fact that models with
sentiment weighting (prefix~$\alpha$) perform well here indicates that
turning points often have an emotional resonance as well as being
suspenseful.

%%%%%%%%%%%%%%%%%%%%%%%%%%%%%%%%%%%%%%%%%%%%%%%%%%%%%%%%%%%%%%%%%%%%%%%%%%%%%%%%
\section{Conclusions}

Our overall findings suggest that by implementing concepts from
psycholinguistic and economic theory, we can predict human judgements
of suspense in storytelling. That uncertainty reduction
($U^\text{Ely}$) outperforms probability-only ($S^\text{Hale}$) and
state-only ($S^\text{Ely}$) surprise suggests that, while
consequential state change is of primary importance for suspense, the
probability distribution over the states is also a necessary
factor. Uncertainty reduction therefore captures the view of suspense
as reducing paths to a desired outcome, with more consequential shifts
as the story progresses \cite{DBLP:conf/aaai/ONeillR14,
  ely2015suspense, perreault2018universal}. This is more in line with
the \citet{smuts2008desire} Desire-Frustration view of suspense, where
uncertainty is secondary.
  
Strong psycholinguistic claims about suspense are difficult to make
due to several weaknesses in our approach, which highlight directions
for future research: the proposed model does not have a higher-level
understanding of event structure; most likely it picks up the textual
cues that accompany dramatic changes in the text. One strand of
further work is therefore analysis: Text could be artificially
manipulated using structural changes, for example by switching the
order of sentences, mixing multiple stories, including a summary at the
beginning that foreshadows the work, masking key suspenseful words, or
paraphrasing. An analogue of this would be adversarial examples used
in computer vision. Additional annotations, such as how certain readers
are about the outcome of the story, may also be helpful in better
understanding the relationship between suspense and
uncertainty. Automated interpretability methods as proposed by
\citet{10.5555/3305890.3306024}, could shed further light on models' predictions.

The recent success of language models in wide-ranging NLP tasks
\cite[e.g.,][]{radford2019language} has shown that language models are
capable of learning semantically rich information implicitly. However,
generating plausible future continuations is an essential part of the
model. In text generation, \citet{Fan2019StrategiesFS} have found that
explicitly incorporating coreference and structured event
representations into generation produces more coherent generated
text. A more sophisticated model would incorporate similar ideas.

Autoregressive models that generate step by step alternatives for
future continuations are computationally impractical for longer
rollouts and are not cognitively plausible. They also differ from the
\citet{ely2015suspense} conception of suspense, which is in terms of
Bayesian beliefs over a longer-term future state, not step by step. There is much recent work (e.g.,
\citet{ha2018worldmodels,gregor2018temporal}), on state-space
approaches that model beliefs as latent states using variational
methods. In principle, these would avoid the brute-force calculation
of a rollout and conceptually, anticipating longer-term states aligns
with theories of suspense. 

Related tasks such as inverting the understanding of suspense to utilise the models in generating more suspenseful stories may also prove fruitful.

This paper is a baseline that demonstrates how modern neural network
models can implicitly represent text meaning and be useful in a
narrative context without recourse to supervision. It provides a
springboard to further interesting applications and research on
suspense in storytelling.

% This work has demonstrated that hierarchical neural language models
% are not only promising for a generation but can also model suspense
% in a way that corresponds to human judgement. It also manifests the
% advantages of applying an unsupervised model combined with models
% from theory rather than relying on large amounts of hand-coded
% training data examples. Future work will look to enrich the model
% with more domain knowledge and structure, extensions to other tasks
% such as using peaks for summarisation, and clustering of the
% embeddings or interpretability type methods could provide useful
% story insights beyond a single output judgement.

\section*{Acknowledgments}

The authors would like to thank the anonymous reviewers, Pinelopi Papalampidi and David Hodges for reviews of the annotation task, the AMT annotators, and Mirella Lapata, Ida Szubert, and Elizabeth Nielson for comments on the paper. Wilmot's work is funded by an EPSRC doctoral training
award.

\bibliography{acl2020}
\bibliographystyle{acl_natbib}

\appendix

\section{Pre-processing}
\label{app:a}

WritingPrompts comes from a public forum of short stories and so is naturally noisy. Story authors often use punctuation in unusual ways to mark out sentences or paragraph boundaries and there are lots of spelling mistakes. Some of these cause problems with the GPT model and in some circumstances can cause it to crash. To improve the quality, sentence demarcations are left as they are from the original WritingPrompts dataset but some sentences are cleaned up and others skipped over. Skipping over is also why there sometimes are gaps in the graph plots as the sentence was ignored during training and inference. The pre-processing steps are as follows. Where substitutions are made rather than ignoring the sentence, the token is replaced by the Spacy \citep{spacy2} POS tag. 

\begin{enumerate}
\item \textbf{English Language:} Some phrases in sentences can be non-English, Whatthelang \citep{joulin2016bag} is used to filter out these sentences.
\item \textbf{Non\-dictionary words:} \textit{PyDictionary} and \textit{PyEnchant} and used to check if each word is  a dictionary word. If not they are replaced.
\item \textbf{Repeating Symbols:} Some author mark out sections by using a string of characters such as *************** or !!!!!!!!!!!!. This can cause the Pytorch GPT implementation to break so repeating characters are replaced with a single one.
\item \textbf{Ignoring sentences:} If after all of these replacements there are not three or more GPT word pieces ignoring the POS replacements then the sentence is skipped. The same processing applies to generating sentences in the inference. Occasionally the generated sentences can be nonsense, so the same criteria are used to exclude them.
\end{enumerate}

%%% not refered to in the text, not necessary -- FK

%\section{Software and Tools}
%\label{app:e}
%
%\paragraph{Models} Models are primarily implemented in AllenNLP \citep{Gardner2017AllenNLP} on Pytorch \citep{paszke2017automatic} with Spacy for text processing \citep{spacy2}, and NLTK \citep{Loper:2002:NNL:1118108.1118117} for some sentiment analysis and agreement metrics. 
%
%\paragraph{Analysis} Other significant packages used include Scikit-Learn \citep{scikit-learn}, SciPy \citep{2019arXiv190710121V}, Pandas \citep{mckinney2010data}, Statsmodels \citep{seabold2010statsmodels} and Dask \citep{matthew_rocklin-proc-scipy-2015} for scale. Figure produced with Plotly.% \citep{plotly}.

\section{Mechanical Turk Written Instructions}
\label{app:b}

\begin{table*}[t!]
\centering
\begin{tabularx}{1.0\textwidth}{ |l|X| }
%\begin{tabular}{@{}lll@{}}
\toprule
\textbf{Annotation} & \textbf{Sentence}                                                                                                                                                                                                                     \\ \midrule
NA                  & Clancy Marguerian, 154, private first class of the 150 + army , sits in his foxhole.                                                                                                                                                  \\
Increase            & Tired cold, wet and hungry, the only thing preventing him from laying down his rifle and walking towards the enemy lines in surrender is the knowledge that however bad he has it here, life as a 50 - 100 POW is surely much worse . \\
Increase            & He's fighting to keep his eyes open and his rifle ready when the mortar shells start landing near him.                                                                                                                                \\
Same                & He hunkers lower.                                                                                                                                                                                                                     \\
Increase            & After a few minutes under the barrage, Marguerian hears hurried footsteps, a grunt, and a thud as a soldier leaps into the foxhole.                                                                                                   \\
Same                & The man's uniform is tan , he must be a 50 - 100 .                                                                                                                                                                                    \\
Big Increase        & The two men snarl and grab at each other , grappling in the small foxhole .                                                                                                                                                            \\
Same                & Abruptly, their faces come together.                                                                                                                                                                                                  \\
Decrease            & ``Clancy?''                                                                                                                                                                                                                           \\
Decrease            & ``Rob?''                                                                                                                                                                                                                              \\
Big Decrease        & Rob Hall, 97, Corporal in the 50 - 100 army grins, as the situation turns from life or death struggle, to a meeting of two college friends.                                                                                           \\
Decrease            & He lets go of Marguerian's collar.                                                                                                                                                                                                    \\
Same                & `` Holy shit Clancy , you're the last person I expected to see here ''                                                                                                                                                                \\
Same                & `` Yeah '' `` Shit man , I didn't think I'd ever see Mr. volunteers every saturday morning at the food shelf' , not after The Reorganization at least ''                                                                              \\
Same                & ``Yeah Rob , it is something isn't it ''                                                                                                                                                                                              \\
Decrease            & `` Man , I'm sorry, I tried to kill you there''.                                                                                                                                                                             \\ \bottomrule
\end{tabularx}
\caption{One of the training annotation examples given to Mechanical Turk workers. The annotation labels are the recommended labels. This is an extract from a validation set WritingPrompts story.}

\label{tab:guide_annotations}
\end{table*}

These are the actual instructions given to the Mechanical Turk Annotators, plus the example in Table~\ref{tab:guide_annotations}:

\paragraph{INSTRUCTIONS} For the first HIT there will be an additional training step to pass. This will take about 5 minutes. After this you will receive a code which you can enter in the code box to bypass the training for subsequent HITS. Other stories are in separate HITS, please search for "Story dramatic tension, reading sentence by sentence" to find them. The training completion code will work for all related HITS.

You will read a short story and for each sentence be asked to assess how the dramatic tension increases, decreases or stays the same. Each story will take an estimated 8-10 minutes. Judge each sentence on how the dramatic tension has changed as felt by the main characters in the story, not what you as a reader feel. Dramatic tension is the excitement or anxiousness over what will happen to the characters next, it is anticipation.

Increasing levels of each of the following increase the level of dramatic tension:
\begin{itemize}
\item \textbf{Uncertainty:} How uncertain are the characters involved about what will happen next? Put yourself in the characters shoes; judge the change in the tension based on how the characters perceive the situation.
\item \textbf{Significance:} How significant are the consequences of what will happen to the central characters of the story?
\end{itemize}
An Example: Take a dramatic moment in a story such as a character that needs to walk along a dangerous cliff path. When the character first realises they will encounter danger the tension will rise, then tension will increase further. Other details such as falling rocks or slips will increase the tension further to a peak. When the cliff edge has been navigated safely the tension will drop. The pattern will be the same with a dramatic event such as a fight, argument, accident, romantic moment, where the tension will rise to a peak and then fall away as the tension is resolved.

You will be presented with one sentence at a time. Once you have read the sentence, you will press one of five keys to judge the increase or decrease in dramatic tension that this sentence caused. You will use five levels (with keyboard shortcuts in brackets):

\begin{itemize}

\item \textbf{Big Decrease (A):} A sudden decrease in dramatic tension of the situation. In the cliff example the person reaching the other side safely.
\item \textbf{Decrease (S):} A slow decrease in the level of tension, a more gradual drop. For example the cliff walker sees an easier route out.
\item \textbf{Same (Space):} Stays at a similar level. In the cliff example an ongoing description of the event.
\item \textbf{Increase (K):} A gradual increase in the tension. Loose rocks fall nearby the cliff walker.
\item \textbf{Big Increase (L):} A more sudden dramatic increase such as an argument. The cliff walker suddenly slips and falls.
\end{itemize}

\paragraph{POST ACTUAL INSTRUCTIONS}

In addition to the suspense annotation. The following review questions were asked:
\begin{itemize}
  \item Please write a summary of the story in one or two sentences.
  \item Do you think the story is interesting or not? And why? One or two sentences.
  \item How interesting is the story? 1--5
\end{itemize}
The main purpose of this was to test if the MTurk Annotators were comprehending the stories and not trying to cheat by skipping over. Some further work through can be done to tie these into the suspense measures and also the WritingPrompts prompts. 

\section{Writing Prompts Examples}
\label{app:c}

The numbers are from the full WritingPrompts test set. Since random sampling was done from these from for evaluation the numbers are not in a contiguous block. There are a couple of nonsense sentences or entirely punctuation sentences. In the model these are excluded in pre-processing but included here to match the sentence segmentation. Also there are some unusual break such as ``should n't'', this is because the word segmentation produced by the Spacy tokenizer. 

\subsection{Story 27}

\begin{figure*}[t]
\includegraphics[width=1.0\textwidth]{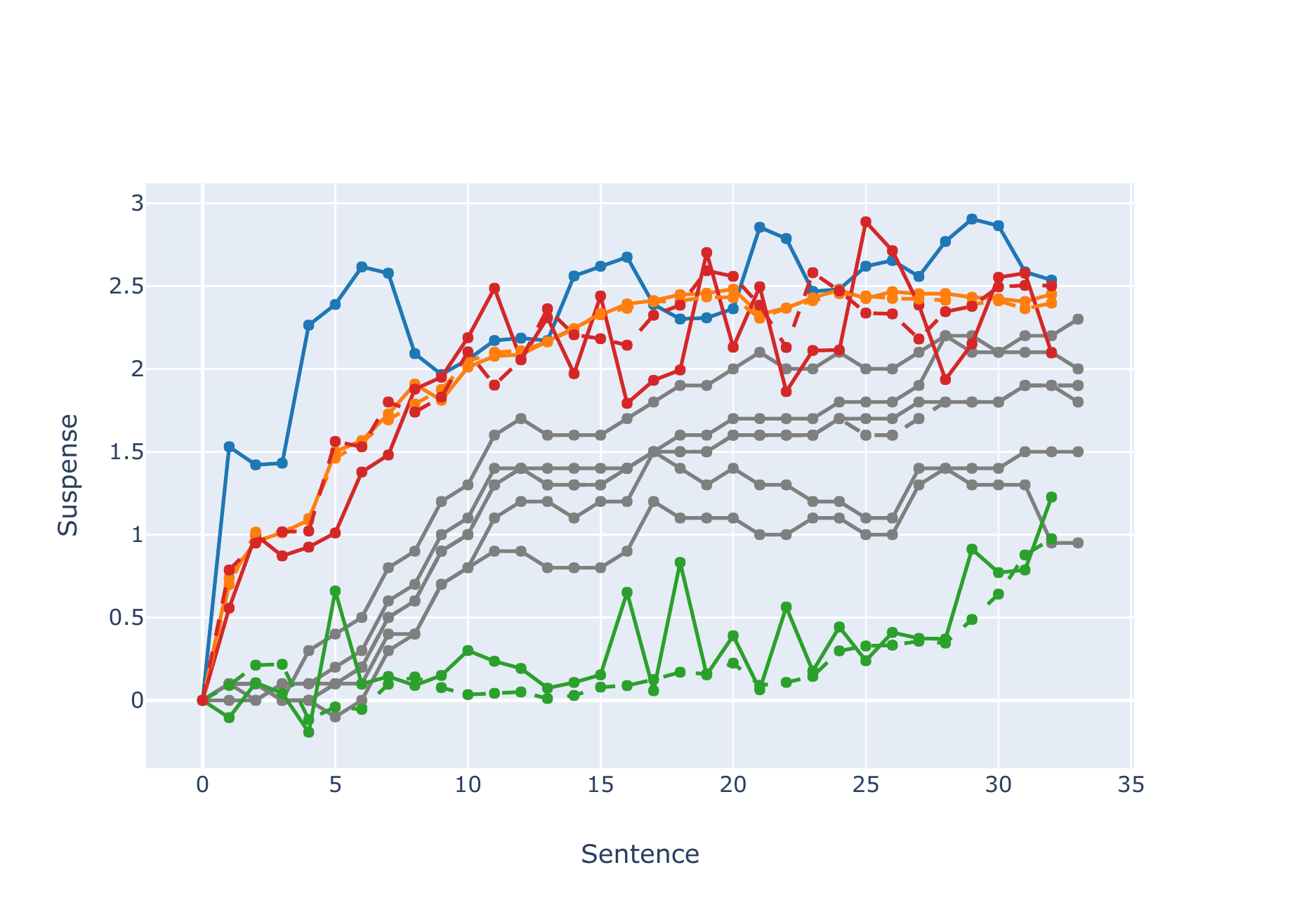}
\caption{Story 27, \textbf{\textcolor{annotatedcolor}{Human}},
\textbf{\textcolor{surpriseentropycolor}{$S^{\text{Hale}}$}}, \textbf{\textcolor{surprisecolor}{$S^{\text{Ely}}$}}, \textbf{\textcolor{suspensecolor}{$U^{\text{Ely}}$}}, \textbf{\textcolor{suspensestatecolor}{$U^{\alpha\text{Ely}}$}}}
\label{fig:wp_27}
\end{figure*}

This is Story 27 from the test set in Figure~\ref{fig:wp_27}, it is the same as the example in the main text:

\begin{enumerate}
\setcounter{enumi}{-1}
\item As I finished up my research on Alligator breeding habits for a story I was tasked with writing , a bell began to ring loudly throughout the office .
\item I could feel the sound vibrating off the cubicle walls .
\item I looked over my cubicle wall to ask a co - worker what the bell was for .
\item I watched as he calmly opened his desk drawer , to reveal a small armory .
\item There were multiple handguns , knives and magazines and other assorted weapons neatly stashed away .
\item `` What the hell is that for ? ''
\item I questioned loudly , and nervously .
\item The man looked me in the eyes , and pointed his handgun at my face .
\item I saw my life flash before my eyes , and could n't understand what circumstances had arisen to put me in this position .
\item I heard the gun fire , and the sound of the shot rang through my ears .
\item I heard something hit the ground loudly behind me .
\item I turned to see the woman who had hired me yesterday , lying in a pool of blood on the floor .
\item She was holding a rifle in her arms .
\item I looked back at the man who had apparently just saved my life .
\item He seemed to be about 40 or so , well built , muscular and had a scar down the right side of his face that went from his forehead down to his beard .
\item `` She liked to go after the new hires '' he explained in a deep voice .
\item `` She hires the ones she wants to kill ''
\item I was n't sure what to make of this , but my thoughts were cut off by the sounds of screaming throughout the building .
\item `` What 's happening ''
\item I asked , barely able to look my savior in the eyes .
\item `` You survive today , and you 'll receive a bonus of \$5,000 and your salary will be raised 5 \% ''
\item I cut the man off .
\item `` What does that ? ''
\item He continued to speak , while motioning me to stop taking .
\item `` I 'll keep you alive , if you give me your bonus and half your raise 
\item He finished .
\item I just nodded , still unable to understand the position I was in .
\item He grabbed my arm so hard I thought it would break , and pulled me over the cubicle wall , and under his desk .
\item Then , he placed a gun in my hand .
\item `` The safety is on , and it 's fully loaded with one in the chamber ''
\item He said , pointing to the safety switch .
\item The weapon felt heavy in my hand , I flicked the safety off with my thumb and gripped the gun tightly .
\item The man looked down at his watch .
\item `` 45 minutes to go ''
\end{enumerate}

\subsection{Story 2066}

\begin{figure*}[t]
\centering
\includegraphics[width=1.0\textwidth]{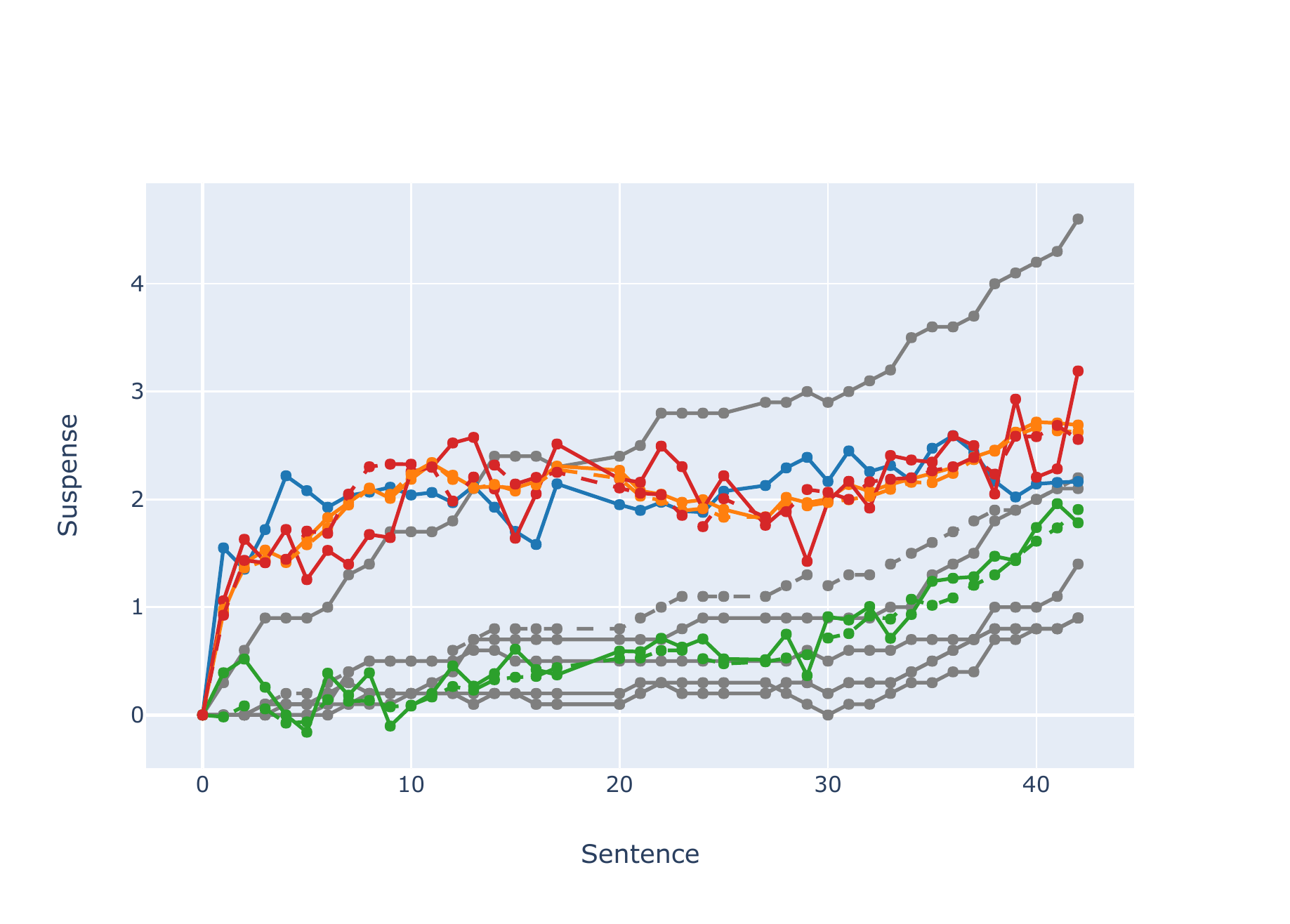}
\caption{Story 2066, \textbf{\textcolor{annotatedcolor}{Human}},
\textbf{\textcolor{surpriseentropycolor}{$S^{\text{Hale}}$}}, \textbf{\textcolor{surprisecolor}{$S^{\text{Ely}}$}}, \textbf{\textcolor{suspensecolor}{$U^{\text{Ely}}$}}, \textbf{\textcolor{suspensestatecolor}{$U^{\alpha\text{Ely}}$}}}
\label{fig:wp_2066}
\end{figure*}

This is Story 2066 from the test set in Figure~\ref{fig:wp_2066}:

\begin{enumerate}
\setcounter{enumi}{-1}
\item The life pods are designed so we ca n't steer .
\item Meant for being stranded in space , it broadcasts an S.O.S .
\item to the entire human empire even as it leaves the mother ship .
\item Within minutes any occupant will be gassed so they wo n't suffer the long months , and perhaps years before a rescue .
\item As soon as your vitals show you 're in deep sleep , it puts the entire interior into a cryogenic freeze .
\item The technology is effective , efficient and brilliant .
\item But as I ' m being launched out of our vessel I ca n't help but slam the hatch with my fists .
\item My ears are still ringing with the endless boom of explosions and my eyes covered in blind spots from the flashes .
\item The battle had been swift , and we humans had lost .
\item Captain 's orders : Abandon ship .
\item Which was why I was stuck here , counting the seconds before I got put into stasis .
\item This was no Titanic .
\item There were ample pods for the entire crew , by the time the call was made only half of us had access to the escape pods , and a quarter of those were injured , a condition that no matter how advanced our technology was , made the life pod a null option .
\item No use being cryogenically frozen if you bleed out before the temperature even drops .
\item Better men and women than I were stuck alive on the ship , and I had to abandon them to whatever their fate may be .
\item I sit back and harness myself into the chair .
\item No use getting worked up over survivor 's guilt now .
\item I 'll do that when I thaw .
\item *
\item * *
\item The first thing I notice is the cold .
\item I ' m too cold .
\item I shiver , my uniform plastered to me .
\item I frown at its tattered appearance .
\item What had happened ?
\item The last thing I remember is ...
\item The life pod .
\item I ' m still in it .
\item But I ' ve been picked up .
\item Someone on the outside has initiated the thaw cycle .
\item At once I ' m struck by relief .
\item Then anxiety .
\item How long was I out ?
\item How many of the crew survived ?
\item Their screams are coming back to me now , and I squirm with the pain .
\item `` Please do n't let me be the only one , '' I whisper to myself , half pleading with fate , half praying to a God .
\item The hatch swings open .
\item The lump in my throat drops to my toes with the weight of lead .
\item A gun greets me .
\item Slowly , I put my hands behind my head .
\item There 's no mistaking the alien wielding it .
\item The brute features are familiar , too familiar .
\item I ' ve been rescued by the wrong side .
\end{enumerate}

\subsection{Story 3203}

\begin{figure*}[t]
\centering
\includegraphics[width=1.0\textwidth]{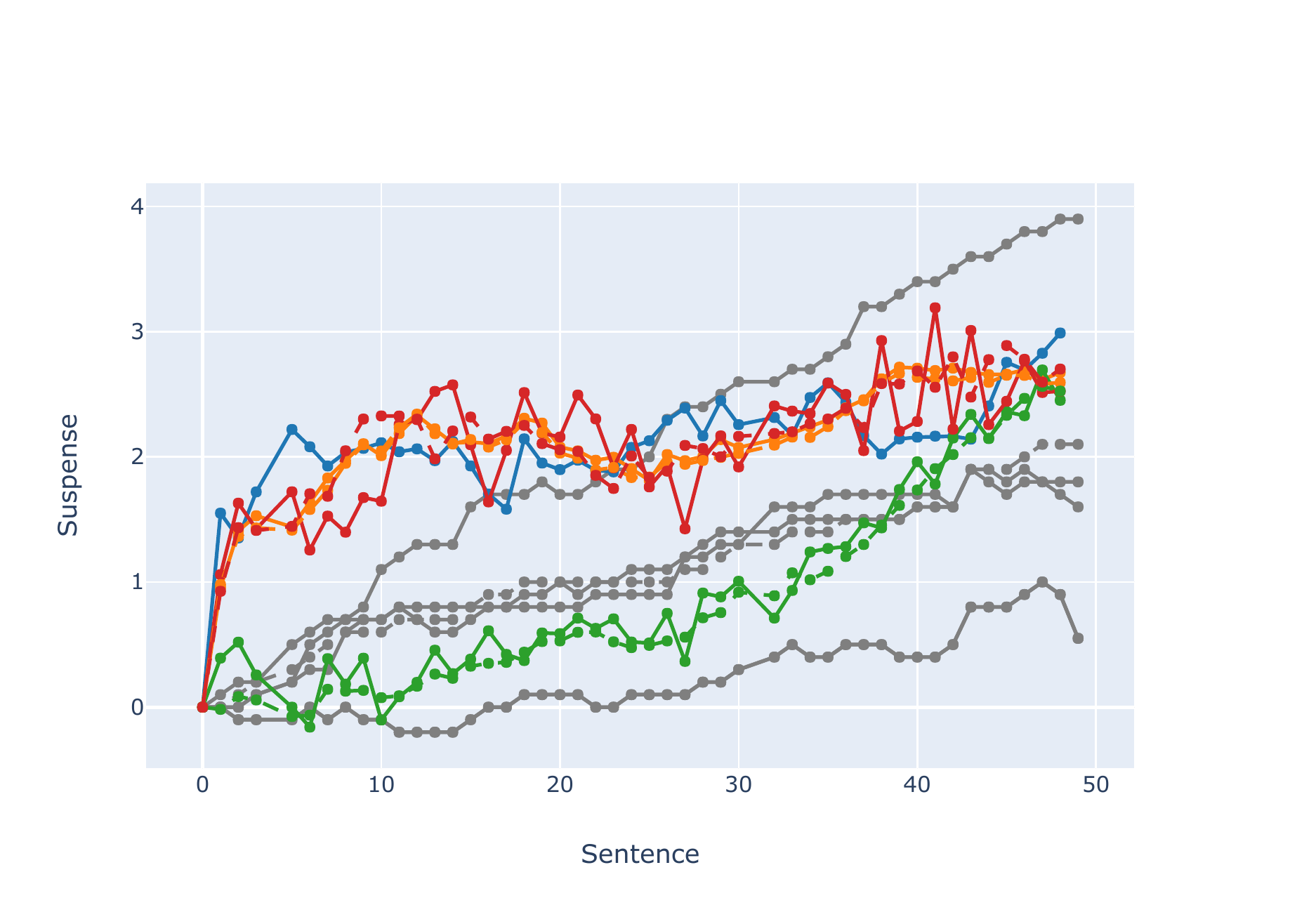}
\caption{Story 3203,  \textbf{\textcolor{annotatedcolor}{Human}},
\textbf{\textcolor{surpriseentropycolor}{$S^{\text{Hale}}$}}, \textbf{\textcolor{surprisecolor}{$S^{\text{Ely}}$}}, \textbf{\textcolor{suspensecolor}{$U^{\text{Ely}}$}}, \textbf{\textcolor{suspensestatecolor}{$U^{\alpha\text{Ely}}$}}}
\label{fig:wp_3203}
\end{figure*}

This is Story 3203 from the test set in Figure~\ref{fig:wp_3203}:

\begin{enumerate}
\setcounter{enumi}{-1}
\item I swore never to kill .
\item I swore that I will never stoop down to their level .
\item That we , the guardians of justice , can and will achieve our goals through the peaceful way .
\item But as I stood there , at the edge of the cliff , staring at the hideous smile that has tormented me for far too long , I could feel my vow slowly breaking before me .
\item `` So what it 's gon na be Batsy ?
\item Will you choose to kill the evil crazy clown , or are you going to let poor Miss Lane fall to her death ?
\item Tick tock tick tock , time 's ticking ! ''
\item I gritted my teeth .
\item Lois was suspended in mid - air , 12 stories high , her life hanging by the mere minutes .
\item Around me , the League lay incapacitated , having fallen to Joker 's devious ambush .
\item I turned towards Clark , hoping that he would have woken up by now .
\item No luck .
\item The Kryptonite knock out gas had worked its miracle .
\item As fate would have it , only two of us are left .
\item Two bitter rivals to the very end .
\item `` Let her go , Joker !
\item This fight is between you and me ! ''
\item I shouted .
\item My mind raced for possible solutions .
\item A well - aimed batarang could free Lois , but I have to rappel to her in time .
\item Too risky with Joker free .
\item I could try knocking him out , but that would not leave me enough time to-
`` Tsk tsk tsk , my dear Bats .
\item Trying to stall for time , are n't you ?
\item How many times must I tell you that it wo n't work !
\item I know you , Bats , better than you know yourself .
\item In fact ... ''
\item He took out a remote , and pushed one of the bright red buttons on it .
\item The cable jerked downwards , closer to the barrel of Joker venom .
\item `` ... for every minute you spend thinking , Miss Lane will be closer to smiley face land .
\item How about that !
\item Hahahaha ! ''
\item It was right then when I lost it .
\item I leaped from my spot , headed straight for the Prince of Clowns .
\item I thought about the last time we almost lost Lois .
\item Clark was so close to unleashing a destructive rampage across Metropolis .
\item Too close .
\item And it was on that day when every member of the League swore an oath to protect Lois no matter what it takes , no matter
what the cost , even if it meant breaking our own sacred vows .
\item Superman was too great an asset to be lost .
\item Joker knew that .
\item From the very moment he saw the destruction Clark unleashed .
\item And he has been targeting Lois ever since .
\item The blade plunged through his chest and into his heart surprisingly quick .
\item I had expected the Joker to have a fail safe mechanism , but apparently he did not .
\item He wanted me to do it .
\item The blood splattered against my suit , as the sickening sound of flesh tearing apart filled my ears .
\item And as all these happened , the Joker kept laughing , his hysterical voice filling the air .
\item He laughed and laughed , until his voice gradually grew weaker , softer .
\item Before he drew his last breath , he raised his bloodied left hand and patted me on my cowl .
\item `` Hehehe ... I win , Batsy . ''
\end{enumerate}

\section{Turning Points Examples}
\label{app:d}

This section is the full text output with some example plots from Turning Points TRIPOD dataset.

\subsection{15 Minutes}

\begin{figure*}[t]
\centering
\includegraphics[width=1.0\textwidth]{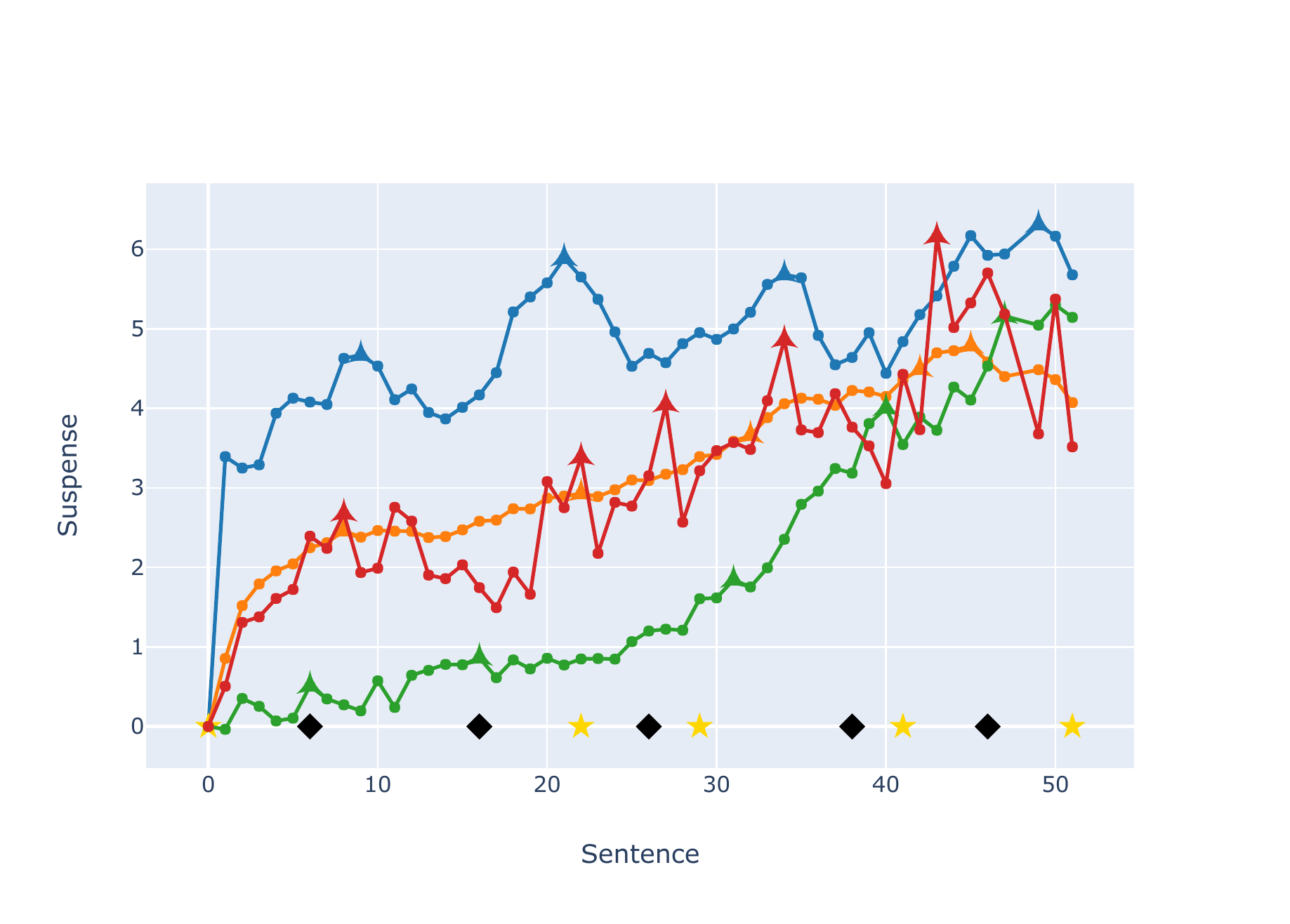}
\caption{The film \textit{15 Minutes},  
\textbf{\textcolor{surpriseentropycolor}{$S^{\text{Hale}}$}}, \textbf{\textcolor{surprisecolor}{$S^{\text{Ely}}$}}, \textbf{\textcolor{suspensecolor}{$U^{\text{Ely}}$}}, \textbf{\textcolor{suspensestatecolor}{$U^{\alpha\text{Ely}}$}},$\medblackdiamond$ theory baseline, {\color{yellow} $\medstar$} TP annotations, triangles are
  predicted TPs.}
\label{fig:tp_train_8}
\end{figure*}

The full text for the synopsis of \href{https://www.imdb.com/title/tt0179626/}{15 Minutes} in Figure~\ref{fig:tp_train_8}, this is the same example as is given in the main text:

\begin{enumerate}
\setcounter{enumi}{-1}
\item After getting out of prison , ex - convicts Emil Slovak ( Karel Roden ) and Oleg Razgul ( Oleg Taktarov ) travel to New York City to meet a contact in order to claim their part of a bank heist in \item Russia ( or somewhere in the Czech Republic ) .
\item Within minutes of arriving , Oleg steals a video camera .
\item They go to the brownstone apartment of their old partner Milos Karlova ( Vladimir Mashkov ) and his wife Tamina , and demand their share .
\item When Milos admits that he spent it , an enraged Emil kills him with a kitchen knife , then breaks Tamina 's neck as Oleg tapes it with his new camera .
\item The couple 's neighbor , Daphne Handlova ( Vera Farmiga ) , witnesses everything , but she escapes before they can get to her .
\item To cover up the crime , they douse the bodies in acetone , carefully position them on the bed , and burn down the apartment , intending to pass it off as an accident .
\item Jordy Warsaw ( Edward Burns ) , an arson investigator , and NYPD detective Eddie Flemming ( Robert De Niro ) are called to the scene .
\item Flemming is a high profile detective who frequently appears on the local tabloid TV show Top Story .
\item Flemming and Warsaw decide to work the case together .
\item They eventually determine that Milos was stabbed so hard that the knife 's tip broke off and lodged in his spine .
\item While checking out the crowd outside , Warsaw spots Daphne trying to get his attention .
\item When he finally gets to where she was , she is gone , but Warsaw manages to produce a sketch of the witness .
\item Emil , who got hold of Daphne 's wallet when she fled the apartment earlier , realizes that Daphne is in the country illegally and will be deported if she calls the police .
\item He contacts an escort service from a business card he found in Daphne 's wallet .
\item He asks for a Czech girl hoping she will arrive .
\item When Honey , a regular call girl , arrives instead , he stabs and kills her , but not before getting the address of the escort service from her .
\item Oleg tapes the entire murder .
\item In fact , he tapes everything he can ; a wannabe filmmaker , he aspires to be the next Frank Capra .
\item Flemming and Warsaw investigate her murder , determine the link to the fire , and also visit the escort service .
\item Rose Heam ( Charlize Theron ) runs the service and tells them that the girl they are looking for ( Daphne ) does not work for her but rather a local hairdresser , and she just told the same thing to \item a couple other guys that were asking the same questions .
\item Flemming and Warsaw then rush to the hairdresser but get there just after Emil and Oleg warn the girl not to say anything to anyone .
\item As Flemming puts Daphne into his squad car , he notices Oleg taping them from across the street .
\item A foot chase begins , culminating in Flemming 's partner getting shot and his wallet stolen .
\item Emil finds a card with Flemming 's name and address in it .
\item He gets very jealous of Flemming 's celebrity status and is convinced that anyone in America can do whatever they want and get away with it .
\item On the night that Flemming is to propose to his girlfriend Nicolette Karas ( Melina Kanakaredes ) , Oleg and Emil sneak into his house and knocks him unconscious , later taping him to a chair .
\item While Oleg is recording , Emil explains his plan - he will kill Flemming , then he will sell the tape to Top Story , and when he is arrested , he will plead insanity .
\item After being committed to an insane asylum he will declare that he is actually sane .
\item Because of double jeopardy , he will get off , collecting the royalties from his books and movies .
\item Flemming starts attacking them with his chair ( while still taped to it ) and almost gets them but Emil stabs him in the abdomen , and putting a pillow on Flemming , killing him .
\item The entire city is in mourning and Emil calls Robert Hawkins ( Kelsey Grammer ) , the host of Top Story , to tell him he has a tape of the killing and is willing to sell it .
\item Robert pays him a million dollars for the tape .
\item Warsaw and the entire police force are furious with Robert and can not believe he would air it , especially since his main reporter is Nicolette .
\item At the same time , Emil and Oleg try to kill Warsaw and Daphne by booby - trapping Daphne 's apartment .
\item The two narrowly escape the resulting fire .
\item On the night it is aired Emil and Oleg sit in a Planet Hollywood to watch it with the rest of the public .
\item As the clip progresses , the customers react with horror at the brutality of it , and a few begin to notice Emil and Oleg are right there with them , Oleg actually smiling at the results of his work , and panic takes place .
\item Emil explains his betrayal to Oleg and as he about to execute Emil with a gun , Oleg stabs him in the arm .
\item The police come in and arrest the wounded Emil , while Oleg escapes .
\item They put Emil in Warsaw 's squad car but instead of taking him to the police station , Warsaw takes him to an abandoned warehouse where he is going to kill him .
\item The police arrive just in time and take Emil away .
\item Everything goes as planned as Emil is now a celebrity and is pleading insanity .
\item His lawyer agrees to work for 30 % of the royalties Emil will receive for his story .
\item Meanwhile , Oleg is jealous of the notoriety that Emil is receiving .
\item While being led away with his lawyer and all the media , Warsaw gets into an argument with the lawyer while the Top Story crew is taping the whole thing .
\item Oleg gives Hawkins the part of the tape where Emil explains his plan to Flemming , proving he was sane the whole time ( Oleg presumably kept this part of the tape on hand as part of an " insurance policy "" ) ."
\item Hawkins shouts out to Emil and explains to him the evidence he now has .
\item Emil pushes a policeman down , takes his gun and shoots Oleg .
\item Emil grabs Flemming 's fiancĂŠe , who is covering the news story , and threatens to shoot her .
\item He is finally cornered by the police and Warsaw .
\item Against orders , Warsaw shoots Emil a dozen times in the chest in order to avenge Eddie 's death .
\item An officer shouts that Oleg is still alive , and Hawkins rushes to him to get footage just as Oleg says the final few words to his movie he is taping just before he dies ( with the Statue of Liberty in the background ) .
\item Shortly afterward , Hawkins approaches Warsaw and tries to cultivate the same sort of arrangement he had with Flemming , suggesting the power an arrangement would give him .
\item In response , Warsaw punches out Hawkins and leaves the scene as the police officers smile in approval .
\end{enumerate}

\subsection{Pretty Woman}

\begin{figure*}[t]
\centering
\includegraphics[width=1.0\textwidth]{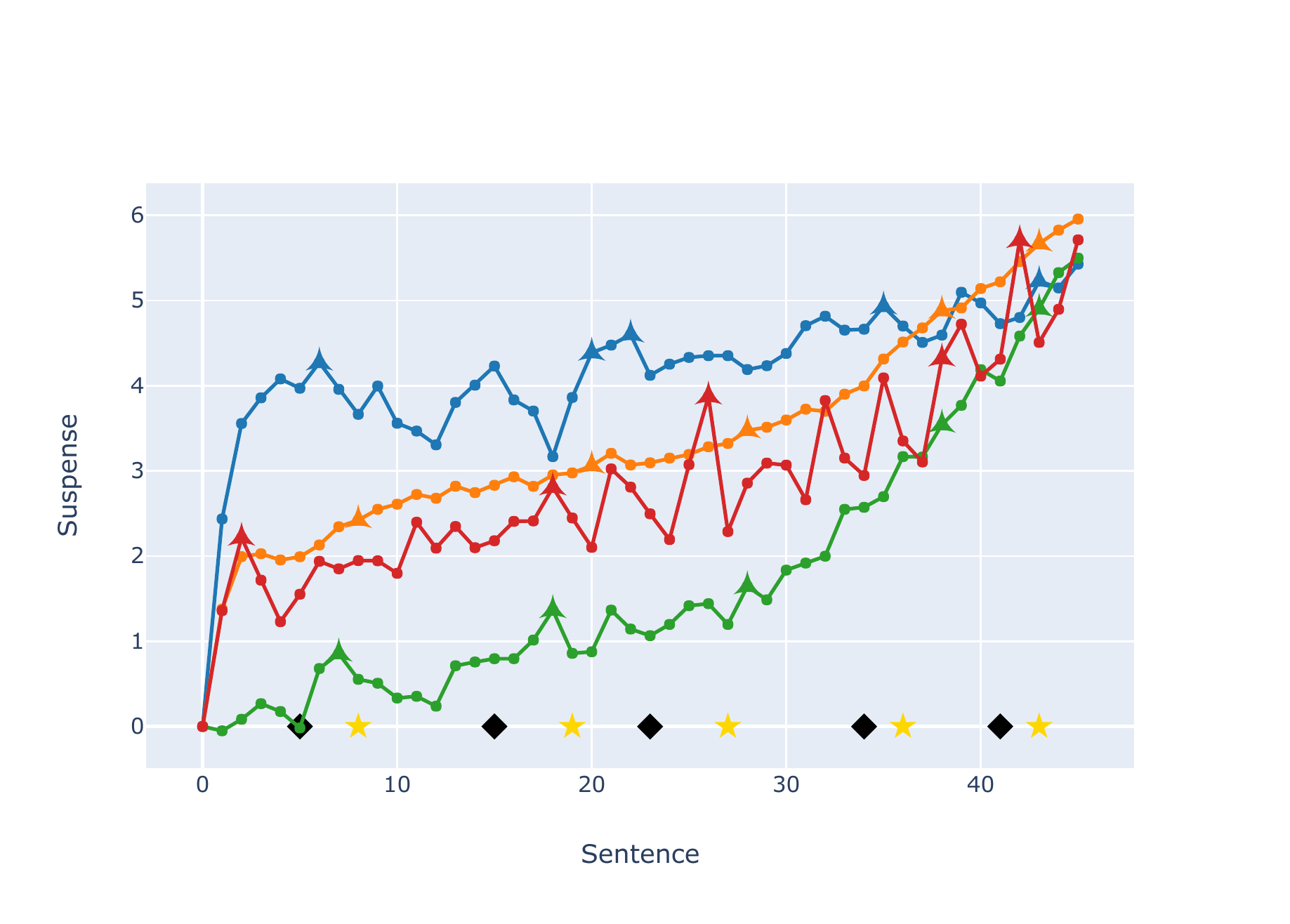}
\caption{The film \href{https://www.imdb.com/title/tt0100405/}{Pretty Woman},
 \textbf{\textcolor{surpriseentropycolor}{$S^{\text{Hale}}$}}, \textbf{\textcolor{surprisecolor}{$S^{\text{Ely}}$}}, \textbf{\textcolor{suspensecolor}{$U^{\text{Ely}}$}}, \textbf{\textcolor{suspensestatecolor}{$U^{\alpha\text{Ely}}$}}, $\medblackdiamond$ theory baseline, {\color{yellow} $\medstar$} TP annotations, triangles are
  predicted TPs.}
\label{fig:tp_train_16}
\end{figure*}

The full text for the synopsis of the film \href{https://www.imdb.com/title/tt0100405/}{Pretty Woman} in Figure~\ref{fig:tp_train_16}: 

\begin{enumerate}
\setcounter{enumi}{-1}
\item Edward Lewis (Gere), a successful businessman and "corporate raider", takes a detour on Hollywood Boulevard to ask for directions.
Receiving little help, he encounters a prostitute named Vivian Ward (Roberts) who is willing to assist him in getting to his destination.

\item The morning after, Edward hires Vivian to stay with him for a week as an escort for social events.
\item Vivian advises him that it "will cost him," and Edward agrees to give her \$3,000 and access to his credit cards.
\item Vivian then goes shopping on Rodeo Drive, only to be snubbed by saleswomen who disdain her because of her unsophisticated appearance.
\item Initially, hotel manager Barnard Thompson (Hector Elizondo) is also somewhat taken aback.
\item But he relents and decides to help her buy a dress, even coaching her on dinner etiquette.
\item Edward returns and is visibly amazed by Vivian's transformation.
The business dinner does not end well, however, with Edward making clear his intention to dismantle Morse's corporation once it was bought, close down the shipyard which Morse spent 40 years building, and sell the land for real estate.
\item Morse and his grandson abandon their dinner in anger, while Edward remains preoccupied with the deal afterward.
\item Back at the hotel, Edward reveals to Vivian that he had not spoken to his recently deceased father for 14 and half years.
\item Later that night, the two make love on the grand piano in the hotel lounge.

\item The next morning, Vivian tells Edward about the snubbing that took place the day before.
\item Edward takes Vivian on a shopping spree.
\item Vivian then returns, carrying all the bags, to the shop that had snubbed her, telling the salesgirls they had made a big mistake.

\item The following day, Edward takes Vivian to a polo match where he is interested in networking for his business deal.
\item While Vivian chats with David Morse, the grandson of the man involved in Edward's latest deal, Philip Stuckey (Edward's attorney) wonders if she is a spy.
\item Edward re-assures him by telling him how they met, and Philip (Jason Alexander) then approaches Vivian and offers to hire her once she is finished with Edward, inadvertently insulting her.
\item When they return to the hotel, she is furious with Edward for telling Phillip about her.
\item She plans to leave, but he apologizes and persuades her to see out the week.
\item Edward leaves work early the next day and takes a breath-taking Vivian on a date to the opera in San Francisco in his private jet.
She is clearly moved by the opera (which is La Traviata, whose plot deals with a rich man tragically falling in love with a courtesan).

\item While playing chess with Edward after returning, Vivian persuades him to take the next day off.
\item They spend the entire day together, and then have sex, in a personal rather than professional way.
\item Just before she falls asleep, Vivian admits that she's in love with Edward.
\item Over breakfast, Edward offers to put Vivian up in an apartment so he can continue seeing her.
\item She feels insulted and says this is not the "fairy tale" she wants.
\item He then goes off to work without resolving the situation.
\item Vivian's friend, Kit De Luca (Laura San Giacomo), comes to the hotel and realizes that Vivian is in love with Edward.

\item Edward meets with Mr. Morse, about to close the deal, and changes his mind at the last minute.
\item His time with Vivian has shown him another way of living and working, taking time off and enjoying activities for which he initially had little time.
\item As a result, his strong interest towards his business is put aside.
\item He decides that he would rather help Morse than take over his company.
\item Furious, Philip goes to the hotel to confront Edward, but only finds Vivian there.
\item He blames her for changing Edward and tries to rape her.
\item Edward arrives in time to stop Philip, chastising him for his greed and ordering him to leave the room.

\item Edward tends to Vivian and tries to persuade her to stay with him because she wants to, not because he's paying her.
\item She refuses once again and returns to the apartment she shares with Kit, preparing to leave for San Francisco to earn a G.E.D.
in the hopes of a better life.
\item Edward gets into the car with the chauffeur that took her home.
\item Instead of going to the airport, he goes to her apartment arriving accompanied by music from La Traviata.
\item He climbs up the fire escape, despite his fear of heights, with a bouquet of roses clutched between his teeth, to woo her.

\item His leaping from the white limousine, and then climbing the outside ladder and steps, is a visual urban metaphor for the knight on white horse rescuing the "princess" from the tower, a childhood fantasy Vivian told him about.
\item The film ends as the two of them kiss on the fire escape.
\end{enumerate}

\subsection{Slumdog Millionaire}

\begin{figure*}[t]
\centering
\includegraphics[width=1.0\textwidth]{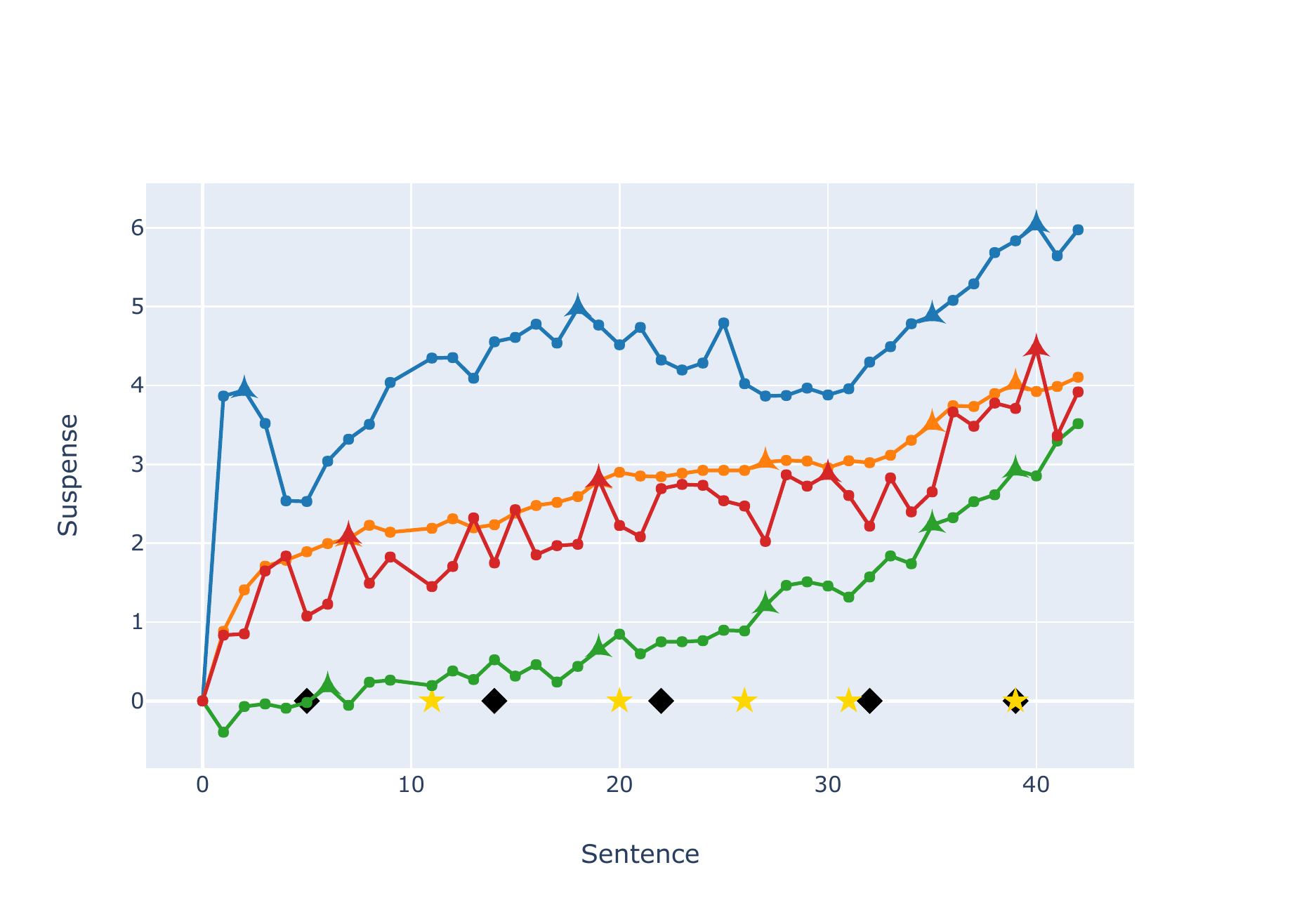}
\caption{\href{https://www.imdb.com/title/tt1010048/}{Slumdog Millionare}, \textbf{\textcolor{surpriseentropycolor}{$S^{\text{Hale}}$}}, \textbf{\textcolor{surprisecolor}{$S^{\text{Ely}}$}}, \textbf{\textcolor{suspensecolor}{$U^{\text{Ely}}$}}, \textbf{\textcolor{suspensestatecolor}{$U^{\alpha\text{Ely}}$}}, $\medblackdiamond$ theory baseline, {\color{yellow} $\medstar$} TP annotations, triangles are
  predicted TPs.}
\label{fig:tp_test_5}
\end{figure*}

The full text for the synopsis of the film \href{https://www.imdb.com/title/tt1010048/}{Slumdog Millionaire}, in Figure~\ref{fig:tp_test_5}: 

\begin{enumerate}
\setcounter{enumi}{-1}
\item In Mumbai in 2006, eighteen-year-old Jamal Malik (Dev Patel), a former street child (child Ayush Mahesh Khedekar, adolescent Tanay Chheda) from the Juhu slum, is a contestant on the Indian version of Who Wants to Be a Millionaire?, and is one question away from the grand prize.
\item However, before the Rs.
\item 20 million question, he is detained and interrogated by the police, who suspect him of cheating because of the impossibility of a simple "slumdog" with very little education knowing all the answers.
\item Jamal recounts, through flashbacks, the incidents in his life which provided him with each answer.
\item These flashbacks tell the story of Jamal, his brother Salim (adult Madhur Mittal, adolescent Ashutosh Lobo Gajiwala, child Azharuddin Mohammed Ismail), and Latika (adult Freida Pinto, adolescent Tanvi Ganesh Lonkar, child Rubina Ali).
\item In each flashback Jamal has a point to remember one person, or song, or different things that lead to the right answer of one of the questions.
\item The row of questions does not correspond chronologically to Jamal's life, so the story switches between different periods (childhood, adolescence) of Jamal.
\item Some questions do not refer to points of his life (cricket champion), but by witness he comes to the right answer.
\item Jamal's flashbacks begin with his managing, at age five, to obtain the autograph of Bollywood star Amitabh Bachchan, which his brother then sells, followed immediately by the death of his mother during the Bombay Riots.
\item As they flee the riot, they run into a child version of the God Rama, Salim and Jamal then meet Latika, another child from their slum.
\item Salim is reluctant to take her in, but Jamal suggests that she could be the third musketeer, a character from the Alexandre Dumas novel (which they had been studying — albeit not very diligently — in school), whose name they do not know.
\item The three are found by Maman (Ankur Vikal), a gangster who tricks and then trains street children into becoming beggars.
\item When Jamal, Salim, and Latika learn Maman is blinding children in order to make them more effective as singing beggars, they flee by jumping onto a departing train.
\item Latika catches up and takes Salim's hand, but Salim purposely lets go, and she is recaptured by the gangsters.
\item Over the next few years, Salim and Jamal make a living travelling on top of trains, selling goods, picking pockets, working as dish washers, and pretending to be tour guides at the Taj Mahal, where they steal people's shoes.
\item At Jamal's insistence, they return to Mumbai to find Latika, discovering from Arvind, one of the singing beggars, that she has been raised by Maman to become a prostitute and that her virginity is expected to fetch a high price.
\item The brothers rescue her, and Salim draws a gun and kills Maman.
\item Salim then manages to get a job with Javed (Mahesh Manjrekar), Maman's rival crime lord.
\item Arriving at their hotel room, Salim orders Jamal to leave him and Latika alone.
\item When Jamal refuses, Salim draws a gun on him, and Jamal leaves after Latika persuades him to go away (presumably so he wouldn't get hurt by Salim).
\item Years later, while working as a tea server at an Indian call centre, Jamal searches the centre's database for Salim and Latika.
\item He fails in finding Latika but succeeds in finding Salim, who is now a high-ranking lieutenant in Javed's organization, and they reunite.
\item Salim is regretful for his past actions and only pleads for forgiveness when Jamal physically attacks him.
\item Jamal then bluffs his way into Javed's residence and reunites with Latika.
\item While Jamal professes his love for her, Latika asks him to forget about her.
\item Jamal promises to wait for her every day at 5 o'clock at the VT station.
\item Latika attempts to rendezvous with him, but she is recaptured by Javed's men, led by Salim.
\item Jamal loses contact with Latika when Javed moves to another house, outside of Mumbai.
\item Knowing that Latika watches it regularly, Jamal attempts to make contact with her again by becoming a contestant on the show Who Wants to Be a Millionaire?
\item He makes it to the final question, despite the hostile attitude of the show's host, Prem Kumar (Anil Kapoor), and becomes a wonder across India.
\item Kumar feeds Jamal the incorrect response to the penultimate question and, when Jamal still gets it right, turns him into the police on suspicion of cheating.
\item Back in the interrogation room, the police inspector (Irrfan Khan) calls Jamal's explanation "bizarrely plausible", but thinks he is not a liar and, ripping up the arrest warrant, allows him to return to the show.
\item At Javed's safehouse, Latika watches the news coverage of Jamal's miraculous run on the show.
\item Salim, in an effort to make amends for his past behaviour, quietly gives Latika his mobile phone and car keys, and asks her to forgive him and to go to Jamal.
\item Latika, though initially reluctant out of fear of Javed, agrees and escapes.
\item Salim fills a bathtub with cash and sits in it, waiting for the death he knows will come when Javed discovers what he has done.
\item Jamal's final question is, by coincidence, the name of the third musketeer in The Three Musketeers, a fact he never learned.
\item Jamal uses his Phone-A-Friend lifeline to call Salim's cell, as it is the only phone number he knows.
\item Latika succeeds in answering the phone just in the nick of time, and, while she does not know the answer, tells Jamal that she is safe.
\item Relieved, Jamal randomly picks Aramis, the right answer, and wins the grand prize.
\item Simultaneously, Javed discovers that Salim has helped Latika escape after he hears Latika on the show.
\item He and his men break down the bathroom door, and Salim kills Javed, before being gunned down himself at the hands of Javed's men.
\item With his dying breath, Salim gasps, "God is great."
\item Later that night, Jamal and Latika meet at the railway station and kiss.
\item The movie ends with a dance scene on the platform to "Jai Ho".
\end{enumerate}

\end{document}